%% file: main.tex
\documentclass[twoside,11pt]{article}

\usepackage{blindtext}

\usepackage{sverl-notation}
\usepackage{subcaption}
\usepackage{svg}
\usepackage{booktabs}
\usepackage{multirow}
\usepackage{adjustbox}

\usepackage[preprint]{jmlr2e}
\usepackage[capitalize,noabbrev]{cleveref}
\creflabelformat{equation}{#2#1#3}

\definecolor{gridred}{HTML}{FFA07A}
\definecolor{gridblue}{HTML}{92C5DE}
\definecolor{gridgreen}{HTML}{B3DE69}

\usepackage{lastpage}
\jmlrheading{}{}{1-\pageref{LastPage}}{4/25}{}{}{Daniel Beechey, Thomas M. S. Smith and \"Ozg\"ur \c{S}im\c{s}ek}

\ShortHeadings{Explaining Reinforcement Learning with Shapley Values}{Beechey, Smith and \c{S}im\c{s}ek}
\firstpageno{1}

\begin{document}

\nocite{Beechey2023}

\title{A Theoretical Framework for Explaining Reinforcement Learning with Shapley Values\thanks{This is an extended version of an earlier conference paper by the same authors: Daniel Beechey, Thomas M. S. Smith and \"Ozg\"ur \c{S}im\c{s}ek. Explaining reinforcement learning with Shapley values. \textit{Proceedings of the 40th International Conference on Machine Learning}, PMLR 202:2003-2014, 2023.}}

\author{\name Daniel Beechey \email djeb20@bath.ac.uk \\
       \name Thomas M. S. Smith \email tmss20@bath.ac.uk \\
       \name \"Ozg\"ur \c{S}im\c{s}ek \email o.simsek@bath.ac.uk \\
       \addr Department of Computer Science \\
       University of Bath \\
       Claverton Down, Bath, BA2 7AY, U.K.}

\editor{}

\maketitle

\input{sections/0_abstract}


\input{sections/1_introduction}

\input{sections/2_background}

\input{sections/3_explaining_rl}

\input{sections/4_explaining_behaviour}

\input{sections/5_explaining_outcomes}

\input{sections/6_explaining_predictions}

\input{sections/7_examples}

\input{sections/8_discussion}

\input{sections/9_related_work}

\input{sections/10_future_work}

\input{sections/11_conclusion}

\input{sections/acknowledgements}


\appendix

\input{sections/appendix/proofs}

\input{sections/appendix/additional_examples}


{
  
\fontsize{11pt} {12.4pt}\selectfont
\bibliography{main}
  
}

\end{document}

%% file: sections/0_abstract.tex
\begin{abstract}
Reinforcement learning agents can achieve super-human performance in complex decision-making tasks, but their behaviour is often difficult to understand and explain. This lack of explanation limits deployment, especially in safety-critical settings where understanding and trust are essential. We identify three core explanatory targets that together provide a comprehensive view of reinforcement learning agents: behaviour, outcomes, and predictions. We develop a unified theoretical framework for explaining these three elements of reinforcement learning agents through the influence of individual features that the agent observes in its environment. We derive feature influences by using Shapley values, which collectively and uniquely satisfy a set of well-motivated axioms for fair and consistent credit assignment. The proposed approach, \emph{Shapley Values for Explaining Reinforcement Learning} (SVERL), provides a single theoretical framework to comprehensively and meaningfully explain reinforcement learning agents. It yields explanations with precise semantics that are not only interpretable but also mathematically justified, enabling us to identify and correct conceptual issues in prior explanations. Through illustrative examples, we show how SVERL produces useful, intuitive explanations of agent behaviour, outcomes, and predictions, which are not apparent from observing agent behaviour alone. 
\end{abstract}

%% file: sections/1_introduction.tex
\section{Introduction}
\label{sec:introduction}

Reinforcement learning agents learn how to act through interaction with their environment, by taking actions, observing their consequences, and receiving scalar reward signals as feedback. They have achieved remarkable performance across diverse domains, including mastering games \citep{Mnih2015, Silver2016, Silver2017, Vinyals2019, Schrittwieser2020, Wurman2022} and controlling complex physical systems \citep{Bellemare2020, Degrave2022, Seo2024}. Despite these successes, reinforcement learning agents remain difficult to interpret: they learn how to act but not how to explain themselves. This lack of explanation hinders deployment, particularly in safety-critical settings where accountability, oversight, and trust are essential. As one example, if a human operator cannot understand an agent's decisions, should they entrust it with control over a nuclear fusion reactor \citep{Degrave2022, Seo2024}?

In this paper, we identify three explanatory targets for reinforcement learning agents that together provide a comprehensive and informative view of the agent. These explanatory targets are behaviour, outcomes, and predictions. \textit{Behaviour} refers to how an agent acts, that is, the decision to take a particular action in a given situation. \textit{Outcomes} refer to the consequences of those actions, for example, the reward collected from the environment. \textit{Predictions} refer to estimates of those outcomes, in other words, any predictions one may want to make about the future interaction of the agent with its environment. Each explanatory target captures a distinct and meaningful aspect of the agent-environment interaction. Together, they enable a comprehensive and structured understanding of the agent.

To explain these three explanatory targets---behaviour, outcomes, and predictions---we propose a unified framework that attributes responsibility to individual \textit{state features}, which are numerical values that describe what the agent observes in its environment. We derive these attributions by using Shapley values \citep{Shapley1953}, which collectively satisfy a set of well-motivated axioms for fair and consistent credit assignment. This approach yields explanations that are not only interpretable but also mathematically justified and uniquely determined by the underlying axioms. We call the proposed framework \emph{Shapley Values for Explaining Reinforcement Learning} (SVERL). Through illustrative examples, we show how SVERL produces intuitive explanations that improve our understanding of agent behaviour, outcomes, and predictions.

Although Shapley values have previously been applied to explain reinforcement learning agents \citep{Rizzo2019, Carbone2020, He2020, Wang2020, Zhang2020, Liessner2021, Lover2021, Remman2021, Schreiber2021, Theumer2022, Zhang2022}, existing work focuses on individual application domains and lacks a theoretical foundation. They often leave ambiguous what is being explained---conflating behaviour, outcomes, and predictions---or failing to distinguish them altogether. We show that without this clarity, explanations can be ambiguous, misleading, or incorrect. Critically, we show that earlier approaches do not explain outcomes, perhaps the most important and distinguishing aspect of reinforcement learning agents compared to explanations of supervised learning. SVERL explicitly distinguishes among explanatory targets and derives each explanation through its own axiomatic framework, ensuring that every attribution is precise, interpretable, and mathematically grounded. We show how existing approaches emerge as special cases of the general theoretical framework developed here, revealing how previous work relates to each other, what assumptions they make, and where their limitations lie.

In addition to Shapley-based methods, previous work has explained reinforcement learning agents by exploiting knowledge about their environment to analyse their learning dynamics \citep{Amir2018, Huang2018, Rupprecht2019, Cruz2019, Juozapaitis2019, Madumal2020, Cruz2023}; using simpler, more interpretable agent models to trade expressivity for transparency \citep{Bastani2018, Jhunjhunwala2019, Silva2020, Topin2021, Bewley2021}; and attributing feature influence with alternative methods that lack the theoretical guarentees of Shapley values \citep{Wang2016, Greydanus2018, Puri2019}. In contrast, SVERL provides explanations that are theoretically grounded and generally applicable, without assumptions about models, dynamics, or environment structure.

While our focus is on establishing the theoretical foundations of SVERL, we also explore some of its natural extensions. Because it inherits the foundations of Shapley values, SVERL can naturally incorporate developments from the broader attribution literature. 
We illustrate this by showing, for example, how its explanations can be approximated when exact computation is impractical, enabling use in real-world settings, and how the framework might support explanations that extend beyond individual states to offer a broader view of agent behaviour, outcomes, and predictions.

These contributions position SVERL as a principled foundation for producing well-defined, interpretable, and comprehensive explanations of reinforcement learning agents.

The remainder of this paper is structured as follows. In \cref{sec:background}, we provide an overview of reinforcement learning and Shapley values. In \cref{sec:erl}, we introduce the motivation and foundational questions behind explaining reinforcement learning agents. In \cref{sec:explainingbehaviour,sec:explainingoutcome,sec:explainingvalueestimation}, we develop a unified framework for explaining agent behaviour, outcomes, and predictions. In \cref{sec:examples}, we present illustrative examples. In \cref{sec:discussion}, we discuss how the framework can be interpreted and used in practice. We review related work in \cref{sec:relatedwork}, propose directions for future research in \cref{sec:futurework}, and conclude in \cref{sec:conclusion}.

%% file: sections/2_background.tex
\section{Background}
\label{sec:background}

In this section, we introduce our notation and briefly describe reinforcement learning \citep{Sutton2018}, Shapley values \citep{Shapley1953} and how Shapley values can be used to explain supervised learning models \citep{Strumbelj2014, Lundberg2017, Frye2020, Covert2021}.

We indicate random variables by capital letters (e.g. $X$, $S$), their values by lowercase letters (e.g. $x$, $s$) and their domains with calligraphic fonts (e.g. $\X$, $\S$). Functions are also denoted by lowercase letters (e.g. $g$); it will be clear from context whether a lowercase letter refers to a function or a value. Random variables are assumed to be vectors, with $\X_i$ indicating the domain for value $x_i$ at index $i$. Given a base set $\F = \{1, 2, \dots, n\}$, where $n$ is the dimension of the domain $\X$, we use $\bar\C\equiv\F\setminus\C$ to represent a set's complement, and $x_\C\equiv \{x_i : i \in \C\}$ to denote a subset of values for $\C\subseteq\F$. Additionally, we use $\Delta(\A)$ to denote the probability simplex over the domain $\A$. That is, the function $\rho : S \rightarrow \Delta(\A)$ expresses a probability mass function $\rho(\cdot|s)$ over $\A$, for each $s \in \S$.

\subsection{Reinforcement Learning}
\label{subsec:rl}

Reinforcement learning is an approach to artificial intelligence that models the interaction of an agent with its environment as the agent seeks to achieve desired outcomes. We follow the common approach of modelling this interaction as a Markov Decision Process (MDP), defined as a 5-tuple $(\S, \A, p, r, \gamma)$. The environment is initiated in state $S_0 \in \S$, according to some initial state distribution $d(s) \defeq \text{Pr}(S_0 = s)$. At decision stage $t$, $t \ge 0$, the agent observes the current state of the environment, $S_t \in \S$, and executes action $A_t \in \A(s_t)$, where $\A(s_t)$ is the set of executable actions in $s_t$. Consequently, the environment transitions to a new state, $S_{t+1} \in \S$, according to the transition kernel $\p(s' | s, a) \defeq \text{Pr}(S_{t+1}=s'\,|\,S_t=s,A_t=a)$, and returns a reward $R_{t+1} \in \mathbb{R}$, whose expectation for a transition from state $s$ to state $s'$ under action $a$ is denoted $r(s,a, s')=\mathbb{E}[R_{t+1}\,|\,S_t=s, A_t=a, S_{t+1}=s']$. The \emph{return} $G_t$ from decision stage $t$ is defined as the discounted sum of all future rewards:
\begin{equation}\nonumber
    G_t = \sum_{k=0}^\infty\gamma^k R_{t+k+1},
\end{equation}
where $\gamma \in [0, 1]$ is a discount factor that weights the relative importance of future rewards.

We assume that states can be decomposed into features, $S=(S_1, S_2, \dots, S_n)$, where each $S_i \in \S_i$ represents the (random) value of an individual feature, such as the velocity of an autonomous vehicle. We will use these state features as the basic elements of our explanations of reinforcement learning.

A policy $\pi: \S \rightarrow \Delta(\A)$ maps each state to a probability distribution over actions. The agent's objective is to learn an \emph{optimal policy} that maximises the expected return from every state. Reinforcement learning methods commonly compute the optimal policy by alternating between policy evaluation: estimating the expected return from state $s$ when following policy $\pi$, $v^\pi(s) \defeq \mathbb{E}_\pi{\left[G_t\,|\,S_t = s\right]}$, or the expected return of executing action $a$, $q^\pi(s,a) \defeq \mathbb{E}_\pi\left[G_t\,|\,S_t = s, A_t = a\right]$, and policy improvement: deriving a new policy by acting greedily with respect to $q^\pi(s,a)$:
\begin{equation}\nonumber
    \underset{a\in\A}{\argmax}\,q^\pi(s,a).
\end{equation}

An elegant and widely-used algorithm for learning the value of the optimal policy is Q-Learning \citep{Watkins1992}. With this algorithm, $q^\pi(s,a)$ is estimated by some representation $Q(s, a)$, which is updated using the temporal-difference update rule below, where $\alpha$ is the step-size parameter:
\begin{equation}\nonumber
    Q(S_t, A_t) \leftarrow Q(S_t, A_t) + \alpha\left(R_{t+1} + \gamma \max_{a \in \A}Q(S_{t+1}, a) - Q(S_t, A_t)\right).
\end{equation}

Reinforcement learning agents can use algorithms like Q-learning to learn an optimal policy. However, they do not explain the reasoning behind their behaviour while learning or executing the policy.

\subsection{Shapley Values in Game Theory}
\label{subsec:sv}

When individuals collaborate to achieve a shared outcome, a fundamental question arises: How should the outcome be fairly divided among them? Cooperative game theory provides a framework for addressing this question by modelling player interactions. This framework produced the \emph{Shapley value}, introduced by \citet{Shapley1953} to fairly assign credit to each participant based on their contribution in different possible coalitions.

Formally, a \emph{coalitional game} consists of a set of players, $\F$, and a characteristic function $v: 2^{|\F|} \to \mathbb{R}$, which assigns a value to each coalition $\C \subseteq \F$, representing the outcome they achieve together. That is, $v(\C)$ denotes the payoff obtained by the subset of players $\C$. By convention, the value assigned to an empty coalition, $v(\emptyset)$, is often set to zero: $v(\emptyset) = 0$ \citep{Vonneumann1947}, though in some cases, a fixed baseline value is used instead \citep{Aas2021}. Coalitional games are sometimes also assumed to satisfy \emph{superadditivity}, meaning that merging coalitions should not decrease their total payoff:
\begin{equation}\nonumber
    v(\C \cup \D) \geq v(\C) + v(\D), \quad \forall \C \subseteq \F, \forall \D \subseteq \F \text{ such that } \C \cap \D = \emptyset.
\end{equation}
We do not assume superadditivity because it is not required for deriving Shapley values and is incompatible with how they are applied in this work.

As an example of a coalitional game, consider a parliament with 100 members, where a vote requires a simple majority to pass; parties A and B both have 49 representatives, and party C has 2 representatives. With each representative given one vote, assumed to agree with their party, we can formulate a vote as a coalitional game played by the three political parties. The outcome of the game is measured by a characteristic function that returns 0 if the vote fails and 1 if the vote passes: $v(\emptyset)=0$, $v(\{\text{A}\}) = 0$, $v(\{\text{B}\}) = 0$, $v(\{\text{C}\}) = 0$, $v(\{\text{A}, \text{B}\}) = 1$, $v(\{\text{A}, \text{C}\}) = 1$, $v(\{\text{B}, \text{C}\}) = 1$ and $v(\{\text{A}, \text{B}, \text{C}\}) = 1$.

A coalitional game allows us to investigate contribution schemes. For example, we might ask, how much does party A contribute to the outcome of a vote relative to parties B and C? Shapley values~\citep{Shapley1953} provide a principled approach to answering such questions by evaluating how the absence of each player affects a game's outcome. The \emph{Shapley value} of player $i \in \F$ in coalitional game $(\F, v)$, denoted by $\phi_i\left(v\right)$, is given by:
\begin{equation}\label{eq:sv}
	\phi_i\left(v\right) = \sum_{\C\subseteq \F\setminus \left\{i\right\}}{\frac{\left|\C\right|!~\left(\left|\F\right|-\left|\C\right|-1\right)!}{\left|\F\right|!} ~ \big(v\left(\C\cup\left\{i\right\}\right)-v\left(\C\right)\big)}.
\end{equation}

Whilst the definition of a coalitional game $(\F, v)$ does not impose an ordering over players, introducing one provides an intuitive interpretation of the Shapley value. Imagine the players in $\F$ joining the game sequentially. When player $i$ joins an existing coalition $\C$, their contribution is the marginal gain in characteristic value: $v\left(\C\cup\left\{i\right\}\right)-v\left(\C\right)$. The Shapley value evaluates this marginal gain for every possible coalition $\C$ that could form without player $i$, weighting each marginal gain by a multinomial coefficient to account for every possible order players in $\C$ may join before player $i$ and players in $\F\setminus\C$ may join after player $i$. Therefore, an equivalent formulation expresses the Shapley value as the expected marginal contribution of $i$ over all possible orderings of players in $\F$:
\begin{equation}\label{eq:svperm}
	\phi_i\left(v\right) = \underset{O \sim \pi(\F)}{\mathbb{E}}\left[v\left(b^i(O) \cup \{i\}\right) - v\left(b^i(O)\right)\right],
\end{equation}
where $\pi(\F)$ is the set of all permutations of $\F$, and $b^i(O)$ represents the set of players that precede $i$ in ordering $O$.

In the parliamentary voting example, the Shapley values for the three parties are: $\phi_{\text{A}} = \frac{1}{3}$, $\phi_{\text{B}} = \frac{1}{3}$, $\phi_{\text{C}} = \frac{1}{3}$. Using the Shapley values, we can conclude that, even though parties A and B both have 49 representatives, and party C has 2 representatives, all three parties make the same contribution towards each vote and hence have the same voting power.

Shapley values are the unique solution to four axioms, described below, that specify the conditions that must be met for a fair allocation of contributions among the players of a coalitional game. While there are other contribution schemes in the literature \citep[such as Banzhaf values;][]{Banzhaf1964}, Shapley values offer the only contribution scheme derived from these axioms of fairness.

Let $\phi_i\left(v\right)$ denote how much player $i$ contributes to the outcome of a coalitional game $(\F, v)$. For fair allocation of credit, the Shapley values $\phi_i\left(v\right)$, $\forall i$, must satisfy:

\begin{axiom}[Efficiency]
    The difference in value between the outcome of the game, $v(\F)$, and the fixed payoff for the game with no players, $v(\emptyset)$, is fully distributed over the set of players:
    \begin{equation}\nonumber
        v(\F) - v(\emptyset) =  \sum_{i \in \F}\phi_i(v).
    \end{equation}
\end{axiom}

\begin{axiom}[Linearity]
    If two independent coalitional games, $u$ and $v$, are combined into a new game with the characteristic function $\alpha u + \beta v$ for some scalars $\alpha, \beta \in \mathbb{R}$, then each player's contribution in the combined game must be the weighted sum of their contributions in the individual games:
    \begin{equation}\nonumber
        \phi_i(\alpha u + \beta v) = \alpha \phi_i(u) + \beta \phi_i(v).
    \end{equation}
\end{axiom}

\begin{axiom}[Symmetry]
    Two players are assigned the same contribution if they make equal marginal contributions to the outcome of the game for all possible coalitions of players:
    \begin{equation}\nonumber
        \phi_i(v) = \phi_j(v) ~~ \text{if} ~~ v(\C \cup \{i\}) = v(\C \cup \{j\}) ~~ \forall \, \C \subseteq \F \setminus \{i, j\}.
    \end{equation}
\end{axiom}

\begin{axiom}[Nullity]
    A player is assigned zero contribution if they make zero marginal contribution to the outcome of the game for all possible coalitions of players:
    \begin{equation}\nonumber
        \phi_i(v) = 0 ~~ \text{if} ~~ v(\C \cup \{i\}) = v(\C) ~~ \forall \, \C \subseteq \F \setminus \{i\}.
    \end{equation}
\end{axiom}

The seminal paper by \citet{Shapley1953} proved that Shapley values are the unique solution that satisfies these four axioms. The efficiency, linearity, and symmetry axioms are sufficient to derive the Shapley value, while the nullity axiom is a direct consequence of them.

\subsection{Explaining Supervised Learning with Shapley Values}
\label{subsec:sv_sl}

Shapley values have been adopted in supervised machine learning to explain model loss \citep{Lipovetsky2001, Owen2014, Lundberg2020, Covert2020, Williamson2020} and model predictions \citep{Strumbelj2009, Strumbelj2010, Strumbelj2011, Strumbelj2014, Datta2016, Lundberg2017}. In this section, we present an overview of the use of Shapley values to explain model predictions. For further details on using Shapley values to explain supervised machine learning, we refer the reader to the unified framework presented by \citet{Covert2021}.

Consider a supervised learning model $f$ that predicts a target variable $Y\in\Y$ for an input $X=(X_1, X_2, \dots, X_n) \in \X$, where $X_i \in \X_i$ represents the value of feature $i$. For example, $f$ could be a model that predicts the rating of a movie using the features genre, duration, and cast.

At the specific input point $x$, the prediction $f(x)$ is made collectively by the feature values $x_1, x_2, \dots, x_n$. This prediction can be modelled as a coalitional game $(\F, f_x)$, where the players $\F$ are the feature values $x$. The outcome of the game is the prediction $f(x)$. The characteristic function is $f_x: 2^{|\F|} \rightarrow \Y$, where $f_x(\C)$ returns the prediction at $x$ when the values of features not in $\C\subseteq\F=\{1, 2, \dots, n\}$ are unknown. The game quantifies how different subsets of feature values $(x_1, x_2, \dots, x_n)$ contribute to the prediction $f(x)$; its Shapley values $\phi_1(f_x), \dots, \phi_n(f_x)$ uniquely satisfy four axioms defining a fair credit allocation, detailed below. When referring to a feature's contribution, we implicitly mean the contribution of its value $x_i$ at input $x$.

\begin{axiom}[Efficiency]
    The difference between the prediction when the values of all features are known, $f_x(\F)$, and the fixed prediction when the values of no features are known, $f_x(\emptyset)$, is fully distributed over the features:
    \begin{equation}\nonumber
        f_x(\F) - f_x(\emptyset) = \sum_{i \in \F}\phi_i(f_x).
    \end{equation}
\end{axiom}

\begin{axiom}[Linearity]
    A feature's contribution to the prediction of a linear sum of independent models, $f$ and $g$, is equal to the sum of the feature's contributions towards the predictions of each model:
    \begin{equation}\nonumber
        \phi_i(\alpha f_x + \beta g_x) = \alpha \phi_i(f_x) + \beta \phi_i(g_x).
    \end{equation}
\end{axiom}

\begin{axiom}[Symmetry]
    Two features that make equal marginal contributions to the prediction of a model, when used (for prediction) with all possible subsets of features, are assigned the same contribution:
    \begin{equation}\nonumber
        \phi_i(f_x) = \phi_j(f_x) ~~ \text{if} ~~ f_x(\C \cup \{i\}) = f_x(\C \cup \{j\}) ~~ \forall \C \subseteq \F \setminus \{i, j\}.
    \end{equation}
\end{axiom}

\begin{axiom}[Nullity]
    A feature that makes zero marginal contribution to the prediction of a model, when used with all possible subsets of features, is assigned zero contribution:
    \begin{equation}\nonumber
        \phi_i(f_x) = 0 ~~ \text{if} ~~ f_x(\C \cup \{i\}) = f_x(\C) ~~ \forall \C \subseteq \F \setminus \{i\}.
    \end{equation}
\end{axiom}

The Shapley values attribute the change in prediction when all feature values are known versus when none are known, $f_x(\F) - f_x(\emptyset)$, to individual feature values. When a prediction $f(x)$ is multidimensional, such as the probability of belonging to different classes in a classification problem, independent Shapley values can be calculated for each element of $f(x)$. 

Computing Shapley values requires making predictions $f_x(\C)$ for all subsets of features $\C\subseteq \F$. However, the function $f(x): \X \rightarrow \Y$ specifies the prediction only when $\C =\F$. It is therefore necessary to make some assumptions on how the model will predict when some features are unknown. There are multiple approaches one can take. One common approach is to represent the unknown feature values by marginalising over their conditional distribution $\p(X_{\bar\C}|X_\C = x_\C)$ \citep{Strumbelj2014, Lundberg2017}. Another common approach is to marginalise over their joint marginal distribution $p(X_{\bar\C})$ \citep{Strumbelj2010}. We briefly explain these two approaches below.

The \emph{conditional characteristic function} represents a model's prediction when feature values $x_{\bar\C}$ are unknown by marginalising out the unknown feature values using their conditional distribution $\p(X_{\bar\C}|X_\C = x_\C)$:
\begin{equation}\nonumber
    f_x(\C) = \mathbb{E}\left[f(X)\,|\,X_\C=x_\C\right].
\end{equation}
An in-depth discussion of the conditional characteristic function is given by \citet{Covert2020} and \citet{Covert2021}, showing that conditional characteristic functions give rise to desirable information-theoretic properties. The \emph{marginal characteristic function}, on the other hand, represents a model's prediction when feature values $x_{\bar\C}$ are unknown by marginalising out the unknown feature values using their joint-marginal distribution $p(X_{\bar\C})$:
\begin{equation}\nonumber
    f_x(\C) = \mathbb{E}\left[f(x_\C, X_{\bar{\C}})\right].
\end{equation}

As a consequence of assuming features are independent, marginal characteristic functions consider impossible feature value combinations when features are dependent \citep{Frye2020}. They were once treated as a relatively easy approximation of conditional characteristic functions. However, it has since been argued that marginal characteristic functions may be a better way to represent unknown feature values because the resulting Shapley value for a feature not explicitly used by a model is always zero; if using the conditional characteristic function, the Shapley value for such a feature may be non-zero \citep{Sundararajan2020, Janzing2020}. 

For example, the function $f(X_1, X_2) = X_1$ never uses the feature $X_2$. However, if $X_2$ depends on $X_1$ (or is correlated with $X_1$), using the conditional characteristic function results in non-zero Shapley values for $x_2$ at any point $x$. Shapley values using the conditional characteristic function ``reflect the model’s dependence on the information communicated by each feature [value], rather than its algebraic dependence'' \citep{Jethani2021}. When using the marginal characteristic function, on the other hand, the Shapley values do not account for any relationship between $x_1$ and $x_2$ because they assume that the features are independent. In other words, with the marginal approach, the Shapley value for $x_2$ is zero: $\phi(x_2)=0$.

In this paper, we mainly use conditional characteristic functions. However, the approach we propose is general and can be used with other ways of defining characteristic functions, including the marginal approach.

\subsection{Global Explanations of Supervised Learning with Shapley Values}
\label{subsec:global_sv}

So far, we have described using Shapley values to provide \emph{local} explanations, which describe how individual feature values contribute to a single prediction $f(x)$. One may also want to understand how feature values influence predictions in general, leading to what are known as \emph{global} explanations. While global Shapley-based explanations of model predictions have not been explored in the literature, similar methods have been used to explain a model’s loss.

Let $\ell(f(x), y)$ denote a loss function that quantifies the discrepancy between the model's prediction $f(x)$ and the target variable $y$. Just as a single prediction can be framed as a coalitional game among feature values, \citet{Covert2020} propose framing a model’s expected loss as a coalitional game $\text{G}=(\F, v^\ell)$, where the feature set $\F$ plays the game, and the characteristic function $v^{\ell}:2^{|\F|}\rightarrow\mathbb{R}$ assigns each subset $\C\subseteq\F$ the expected loss when the features outside $\C$ are unknown:
\begin{equation}\nonumber
    v^{\ell}(\C) = \mathbb{E}_{p(x)}\left[\ell\left(f_x(\C), y\right)\right],
\end{equation}
which is an expectation of a per-instance characteristic function:
\begin{equation}\nonumber
    v^{\ell}_x(\C) \coloneqq \ell\left(f_x(\C), y\right).       
\end{equation}
In this formulation, the global Shapley value of feature $i$, denoted $\phi_i(v^\ell)$, represents its contribution to the overall prediction accuracy of the model. The characteristic function is a linear sum of per-instance characteristic functions $v^{\ell}_x(\C)$ corresponding to separate coalitional games at each input $x$. Therefore, the linearity axiom ensures that the global Shapley values are also an expectation over the Shapley values for those coalitional games:
\begin{equation}\nonumber
    \phi_i(v^{\ell}) = \mathbb{E}_{p(x)}\left[\phi_i(v^{\ell}_x)\right].
\end{equation}
This interpretation makes clear that a global Shapley value reflects the mean contribution of a feature to the model’s loss over the data distribution $p(x)$, providing insight into its overall importance across the dataset.

\subsection{Computing Shapley Values in Practice}
\label{subsec:approximating_sv}

When the number of features or the domain $\X$ is large, the exact computation of Shapley values becomes intractable. In such cases, Shapley values and characteristic functions can be approximated. The conditional characteristic function can be approximated through Monte Carlo sampling \citep{Strumbelj2010, Strumbelj2014}: 
\begin{equation}\label{eq:approxcharateristic}
	f_\C(x) = \mathbb{E}\left[f(X)\,|\,X_\C=x_\C\right] = \lim_{n\rightarrow \infty}{\frac{1}{n}\sum_{x' \,\sim\, \p(\cdot|X_\C = x_\C)}{f(x')}},
\end{equation}
where the conditional distribution $\p(X|X_\C = x_\C)$ can be approximated by a method such as variational inference \citep{Frye2020}.

Two common methods use \cref{eq:approxcharateristic} to approximate Shapley values in supervised learning. The first method combines \cref{eq:approxcharateristic} and the Shapley values of \cref{eq:svperm} into a single Monte Carlo sampling method \citep{Strumbelj2010, Strumbelj2014}:
\begin{equation}\label{eq:approxshapleyvalues}
	\phi_i(f_x) = \lim_{n\rightarrow \infty}{\frac{1}{n}\sum_{\substack{x' \,\sim\, \p(\cdot|X_{\C\cup\{i\}}=x_{\C\cup\{i\}}) \\\\ x'' \,\sim\, \p(\cdot|X_\C = x_\C)}}{f(x')-f(x'')}},
\end{equation}
where $\C$ is the set of features that precede $i$ in ordering $O$ and $O$ is sampled uniformly from $\pi(\F)$, the set of all permutations of $\F$. The second method is the kernel method, which approximates characteristic values using \cref{eq:approxcharateristic}, rewrites the Shapley values of \cref{eq:sv} as the solution to a particular weighted least squares problem, and then uses the approximate characteristic function to solve for the Shapley values of all features at a point $x$ simultaneously. The kernel method was introduced by \citet{Lundberg2017}; we direct the reader to \citet{Aas2021} for a detailed description of the approach.

This concludes our introduction to the relevant background. In the following section, we consider what aspects of agent-environment interaction can and should be explained (\cref{sec:erl}). We then formalise these explanations using Shapley values (\cref{sec:explainingbehaviour,sec:explainingoutcome,sec:explainingvalueestimation}).

%% file: sections/3_explaining_rl.tex
\section{Explaining Reinforcement Learning Agents} 
\label{sec:erl}

We start our exploration by asking \textit{what would be useful to explain} regarding reinforcement learning agents \textit{for people who develop, test, and deploy such agents}. We propose three explanatory targets: behaviour, outcomes, and predictions.

\begin{description}
    \item \textbf{Behaviour} refers to how the agent acts, that is, the decision to take a particular action in a given situation.
    \item \textbf{Outcomes} refer to measurable consequences of an agent's behaviour, such as the next state, immediate reward, or expected return. Here we focus on expected return as the outcome of interest. The explanation framework we develop can be extended to other outcomes. 
    \item \textbf{Predictions} refer to estimates of outcomes. These predictions may be made by the agent itself or by an outside observer, by using information which may or may not match the information used by the agent when deciding how to act.
\end{description}

Each of these explanatory targets---behaviour, outcomes, and predictions---provides a useful perspective on the agent. Together, they help build a comprehensive understanding of an agent's decisions, the consequences of those decisions, and estimates of those consequences.

\begin{figure}
    \centering
    \includegraphics[width=0.5\linewidth]{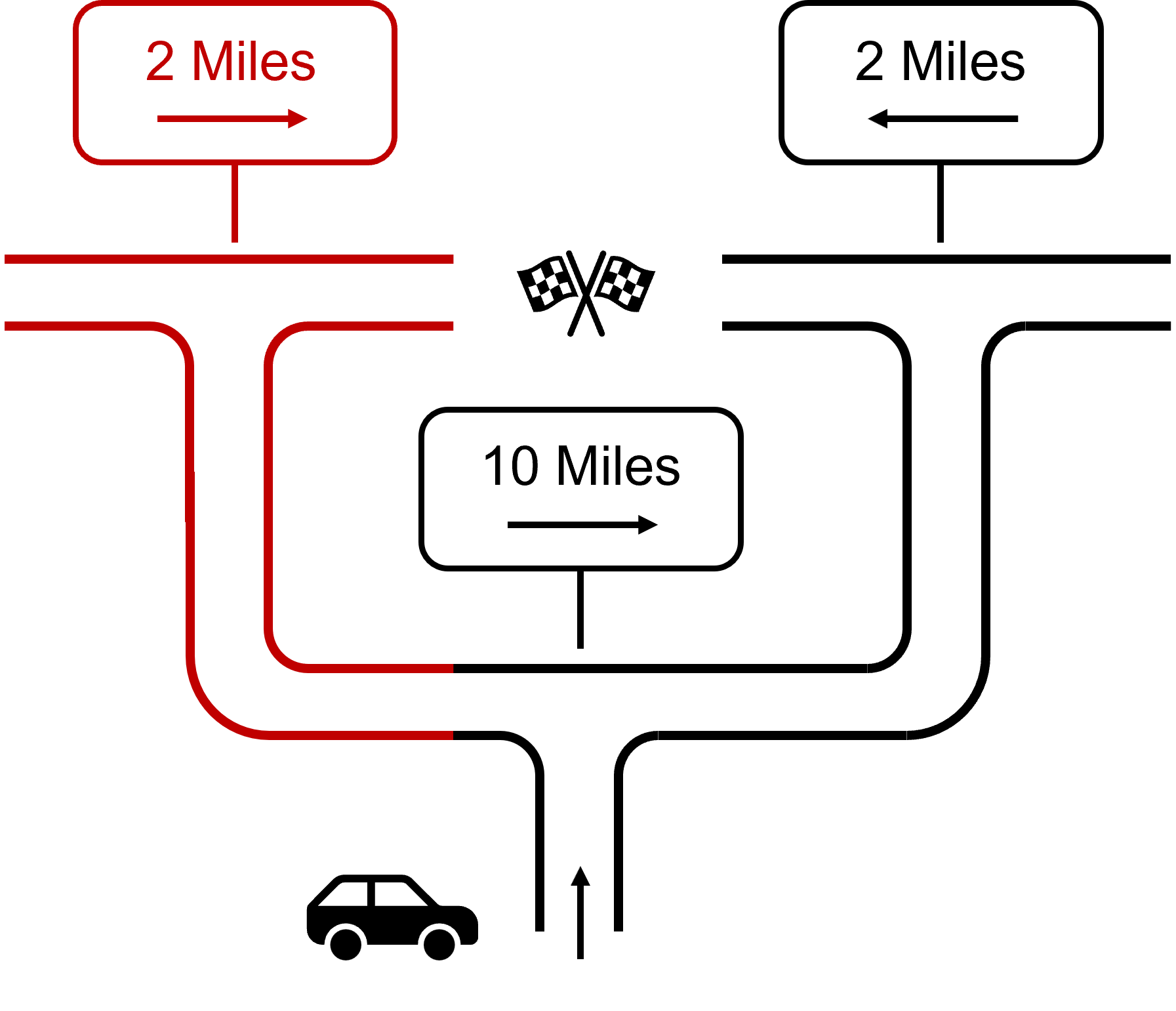} 
    \caption{An autonomous vehicle navigates toward a destination marked by flags. The agent takes the path shown in black ink, turning right at the first junction. The path shown in red ink represents an alternative shortest route.}
    \label{fig:roadsigns}
\end{figure}

Consider an illustrative example depicted in \cref{fig:roadsigns}. An autonomous vehicle is navigating a city using road signs that display, at each junction, the direction and distance to the vehicle's destination. The vehicle follows the directions shown on the road signs. For example, it turns right at the first junction. Consequently, the agent takes the path shown in black ink. An alternative shortest route is shown in red ink.

\textbf{Explaining behaviour} means understanding and describing why an agent acts the way it does, for example, why the vehicle turns right at the first junction.

\textbf{Explaining outcomes} means explaining direct, measurable consequences of an agent's behaviour. For instance, in our example, we may want to explain the total distance travelled by the vehicle. Outcomes are distinct from behaviour, and a change in behaviour does not necessarily result in a change in outcomes. For instance, in our example, at the first junction, turning right or left both lead to the destination via routes of equal length. Consequently, while the road sign at the first junction may influence how the vehicle acts, it does not influence the distance travelled by the vehicle to reach its destination. 


\textbf{Explaining predictions} means explaining an estimate of an outcome. This differs from explaining outcomes themselves in a subtle but important way. In explaining outcomes, we explain the indirect influence of information on an outcome through its influence on behaviour. For example, different information may cause the agent to act differently, which may then lead the agent to collect a different return. In contrast, in explaining predictions, we explain the direct influence information has on predictions of outcomes, with agent behaviour not impacted. In our example, sign directions guide the agent’s decisions and thus shape its outcomes, while sign distances---though not used by the agent---can inform external predictions of those outcomes, such as predicting the agent's expected return based on its distance from the goal. The distinction between explaining outcomes and explaining predictions will become clear throughout the paper and is illustrated in detail in \cref{subsec:contrast}.

By explaining behaviour, outcomes, and predictions, we gain a holistic understanding of a reinforcement learning agent: how it behaves, what its behaviour produces, and how we estimate those results. In the following sections, we formalise these explanations using Shapley values. Specifically, we analyse how to explain behaviour (\cref{sec:explainingbehaviour}), outcomes (\cref{sec:explainingoutcome}), and predictions (\cref{sec:explainingvalueestimation}), introducing a unified framework: Shapley Values for Explaining Reinforcement Learning (SVERL).

%% file: sections/4_explaining_behaviour.tex
\section{Explaining Agent Behaviour}
\label{sec:explainingbehaviour}

To explain agent behaviour, we start by recognising that behaviour is determined by a policy: a function $\pi: \S \rightarrow \Delta(\A)$ that maps each state to a probability distribution over actions. We consider first discrete action spaces, before extending to continuous actions in \cref{subsec:continuousactions}. We assume that each state $s \in \S$ can be decomposed into the values of its features, $s=(s_1, s_2, \dots, s_n)$. This common assumption is generally satisfied in practice. For example, an agent’s location can be represented by its $x$ and $y$ coordinates. Consequently, a sound approach to explaining agent behaviour is to analyse how a policy uses state features to determine $\pi(s, a) \in [0, 1]$, the probability of selecting each action $a$ in state $s$, for all possible values of $s$ and $a$. For example, how does an agent’s $x-$coordinate influence the probability that it moves in a certain direction?

In the following sections, we present a theoretical analysis of explaining the behaviour of reinforcement learning agents using the influence of feature values on the probabilities of selecting actions.

\subsection{State Features Collaborate to Determine Action Probabilities}
\label{subsec:policyshapley}

Having established that agent behaviour can be understood through how feature values influence the probability of selecting actions, we now take a game-theoretic perspective on this relationship. Specifically, we observe that $\pi(s, a)$ can be viewed as the outcome of a coalitional game played by the features at state $s$, which collaborate to determine the probability of selecting action $a$. This naturally leads to the question: how much does each individual feature value contribute to the precise value of $\pi(s, a)$? To answer this, we define fairness axioms that any attribution method should satisfy and identify Shapley values as the unique solution. Below, we formally define the coalitional game, introduce the fairness axioms, and derive the unique solution from first principles.

\begin{definition}[Discrete Policy Game]
    A set $\F=\{1, \dots, n\}$ of features and a characteristic function $\tilde\pi_s^a: 2^{|\F|} \rightarrow \mathbb{R}$, where $\tilde\pi_s^a(\C)$ returns the probability of selecting action $a$ in state $s$ when the values of features $\C \subseteq \F$ are known by the agent and the features $\bar\C = \F \setminus \C$ are unknown.
\end{definition}

Central to the definition of the discrete policy game $(\F, \tilde\pi_s^a)$ is the characteristic function $\tilde\pi_s^a: 2^{|\F|} \rightarrow [0, 1]$, which specifies how the agent would act---that is, the probability of selecting action $a$ when in state $s$---when only a subset of the feature values is known. This function depends on the underlying policy $\pi$, which governs behaviour when all state features are fully observed. However, the policy $\pi$ does not define the agent's behaviour when some feature values are unknown. To account for this, we introduce the characteristic function $\tilde\pi_s^a$ to extend $\pi$ to settings where the agent has partial knowledge of the state. We use the same symbol $\pi$ to highlight this relationship: in the case where all features are known, that is, $\C = \F$, we require that $\tilde\pi_s^a(\C) = \pi(s, a)$. The precise construction of the characteristic function $\tilde\pi_s^a$ will be addressed later in \cref{subsec:policycharacteristic}. For now, we assume it is given and turn to the central question of allocating credit among the features.

The question we now face is: What is the precise contribution of each state feature value $s_i \in \mathcal{S}_i$ to the probability of selecting action $a$ in state $s$? We denote this contribution by $\phi_i(\tilde\pi_s^a)$. In other words, how can we fairly distribute---across the features---the difference between the action probability when all feature values are known and the action probability when none are known? Below, we define four axioms that any attribution $\phi_i(\tilde\pi_s^a)$, for all $i \in \mathcal{F}$, should satisfy to qualify as a fair credit allocation. When referring to a feature's contribution, we implicitly mean the contribution of its value $s_i$ at state $s$.

\begin{axiom}[Efficiency]
    The difference in value between the action probability when all features are known, $\tilde\pi_s^a(\F)$, and the action probability when no features are known, $\tilde\pi_s^a(\emptyset)$, is fully distributed over the features $\F$:
    \begin{equation}\nonumber
        \tilde\pi_s^a(\F) - \tilde\pi_s^a(\emptyset) = \sum_{i \in \F}\phi_i(\tilde\pi_s^a).
    \end{equation}
\end{axiom}

\begin{axiom}[Linearity]
    For a composite policy $\alpha\,\pi + (1- \alpha)\,\omega$ that follows, at each decision stage, policy $\pi$ with probability $\alpha$ and policy $\omega$ with probability $(1- \alpha)$, a state feature's contribution to the action probability $\alpha\,\pi(s, a) + (1- \alpha)\,\omega(s, a)$ is equal to the weighted sum of its contributions to the individual action probabilities $\pi(s, a)$ and $\omega(s, a)$:
    \begin{equation}\nonumber
        \phi_i(\alpha\,\tilde\pi_s^a + (1- \alpha)\,\tilde\omega_s^a) = \alpha\,\phi_i(\tilde\pi_s^a) + (1- \alpha)\,\phi_i(\tilde\omega_s^a).
    \end{equation}
\end{axiom}

\begin{axiom}[Symmetry]
    Two features are assigned the same contribution to the action probability $\pi(s, a)$ if they make equal marginal contributions to the probability of selecting action $a$ in state $s$ for all possible coalitions of features:
    \begin{equation}\nonumber
        \phi_i(\tilde\pi_s^a) = \phi_j(\tilde\pi_s^a) ~~ \text{if} ~~ \tilde\pi_s^a(\C \cup \{i\}) = \tilde\pi_s^a(\C \cup \{j\}) ~~ \forall \C \subseteq \F \setminus \{i, j\}.
    \end{equation}
\end{axiom}
    
\begin{axiom}[Nullity]
    A feature is assigned zero contribution to the action probability $\pi(s, a)$ if it makes zero marginal contribution to the probability of selecting action $a$ in state $s$ for all possible coalitions of features:
        \begin{equation}\nonumber
            \phi_i(\tilde\pi_s^a) = 0 ~~ \text{if} ~~ \tilde\pi_s^a(\C\cup\{i\}) = \tilde\pi_s^a(\C) ~~ \forall \C \subseteq \F \setminus \{i\}.
        \end{equation}
\end{axiom}

These four axioms are directly analogous to the standard axioms from cooperative game theory used to fairly assign player contributions to the outcome of a coalitional game, as discussed in \cref{subsec:sv}.

\begin{proposition}\label{prop:policyshapley}
    The Shapley values $\phi_1(\tilde\pi_s^a), \dots, \phi_n(\tilde\pi_s^a)$ for the discrete policy game $(\F, \tilde\pi_s^a)$ are the unique allocation of the action probability $\pi(s, a)$ across the features of state $s$ that satisfies the four axioms of fair credit assignment: efficiency, linearity, symmetry, and nullity.
\end{proposition}

The proofs for all propositions are presented in \cref{sec:proofs}. The Shapley values of the discrete policy game provide a principled way to attribute action probabilities to individual state features, offering a concrete explanation of agent behaviour.

We now revisit our autonomous vehicle example introduced in \cref{sec:erl,fig:roadsigns}. At the first junction, the vehicle turns right. How can we explain this behaviour? Before we can apply our framework to attribute the decision to the observed features at that state, we must first consider how the agent might behave when one or more features are unknown.

We examine two possible ways the agent could act at the first junction when features are unknown. For clarity, we refer to these behaviours as separate agents. \texttt{Agent A} bases its decisions solely on the direction the sign indicates; if the direction is visible, it follows it. Otherwise, it chooses randomly. \texttt{Agent B} uses both direction and distance to determine which way to turn; if it observes a distance of 10 miles or a direction of right, it moves right. Otherwise, it also chooses randomly. These assumptions give rise to different policies under unknown information, shown in \cref{tab:policy_assumptions}.

\begin{table}[h]
\centering
\begin{tabular}{lcccc}
\toprule
\textbf{Agent} & \textbf{Both features} & \textbf{Direction only} & \textbf{Distance only} & \textbf{No features} \\
\midrule
\texttt{A} & $\pi(\text{L}) = 0,\ \pi(\text{R}) = 1$ & $0,\ 1$ & $0.5,\ 0.5$ & $0.5,\ 0.5$ \\
\texttt{B} & $0,\ 1$ & $0,\ 1$ & $0,\ 1$ & $0.5,\ 0.5$ \\
\bottomrule
\end{tabular}
\caption{Agent behaviour at the first junction under different subsets of observed features. Each cell shows the probability of turning left (L) and right (R).}
\label{tab:policy_assumptions}
\end{table}

We then compute the Shapley values for each policy, attributing the agent’s choice of action to the direction and distance features. The results are shown in \cref{tab:shapley_example}. For \texttt{Agent A}, all credit is assigned to the direction feature since the policy disregards distance entirely. For \texttt{Agent B}, both features contribute equally. In both cases, the features increasing the probability of turning right must decrease the probability of turning left, resulting in negative Shapley values for turning left. This example illustrates how the explanation depends critically on what is assumed about agent behaviour under partial information. We next discuss how we might approach this problem in general.

\begin{table}[h]
\centering
\begin{tabular}{lcccc}
\toprule
\textbf{Agent} & \textbf{Action} & $\phi_{\text{direction}}$ & $\phi_{\text{distance}}$ \\
\midrule
\multirow{2}{*}{\texttt{A}} & L & $-0.5$ & $0$ \\
                            & R & $0.5 = \tfrac{1}{2}[(1 - 0.5) + (1 - 0.5)]$ & $0$ \\
\midrule
\multirow{2}{*}{\texttt{B}} & L & $-0.25$ & $-0.25$ \\
                            & R & $0.25$ & $0.25$ \\
\bottomrule
\end{tabular}
\caption{Shapley values at the first junction for each feature under the two agent policies.}
\label{tab:shapley_example}
\end{table}

\subsection{What Can Be Assumed About Agent Behaviour When Some State Features Are Unknown?}
\label{subsec:policycharacteristic}

We now consider how an agent behaves when some state features are unknown---a question that must be addressed to compute Shapley values. Several approaches might be considered. A natural proposal is to define new learning problems over subsets of observed features---retaining the original environment’s dynamics and rewards---and train policies using reinforcement learning. However, such policies may fail to converge due to partial observability. Even when they do, if the original policy was suboptimal---or not learnt via rewards at all---the newly learned policy may not meaningfully correspond to how the original agent would behave with unknown features, since it solves a different optimisation problem. Another proposal is introducing a special feature value to indicate that a feature is unknown, possibly implemented by replacing unknown features with a fixed placeholder in the agent’s input. But without adapting training, the agent has no reason to interpret such values as indicating missingness. These approaches both require changes to the MDP or training process, limiting generality and making them unsuitable for post hoc explanation---though they may offer promising directions for future work. In what follows, we introduce a general and post hoc approach to evaluating behaviour under unknown feature values.

\begin{definition}
    Let $\mu: \S_\C \rightarrow \Delta(\A)$ be a policy defined over a subset of features $\C$ such that the probability of selecting action $a$ when the values of features $\bar\C$ are unknown is the expected probability of selecting $a$ given the known feature values $s_{\C}$:
    \begin{equation}\nonumber
        \mu(s_\C, a) = \mathbb{E}[\pi(S, a)\,|\,S_\C = s_\C] = \sum_{s' \in \S^+} p^\pi(s' \mid s_\C)\pi(s', a),
    \end{equation}
    where $s_\C$ denotes the values of the features in $\C$ at state $s$. Here, $\S^+$ is the set of non-terminal states. The distribution $p^\pi(S \mid S_\C = s_\C)$ is the conditional steady-state distribution: the probability that an agent following policy $\pi$ is in state $s$, conditioned on observing $s_\C$.
\end{definition}

Among all policies defined over the feature subset $\C$, the policy $\mu$ is the one that deviates the least from $\pi$ in expectation when comparing action probabilities.

\begin{proposition}\label{prop:policychar}
    When comparing policies using mean squared error between the probabilities of selecting action $a$, the policy $\mu$ is the policy that deviates the least in expectation from the policy $\pi$:
    \begin{equation}\label{eq:mu}
        \mu = \argmin_g \mathbb{E}\left[\lVert \pi(S, a) - g(S_\C, a) \rVert^2\right],
    \end{equation}
    where $g: \S_\C \rightarrow \Delta(\A)$ is also defined over the subset of features $\C$.
\end{proposition}

This motivates our definition of the characteristic function $\tilde\pi_s^a(\C)$: the probability of selecting action $a$ at state $s$ when only the features in $\C$ are known is taken to be the value given by the policy $\mu$ at that state.

\begin{definition}[Discrete Policy Characteristic Function]
    The probability of selecting action $a$ when the values of features $\bar\C$ are unknown is defined as the expected probability of selecting $a$ given the known feature values $s_{\C}$:
    \begin{equation}\nonumber
        \tilde\pi_s^a(\C) \defeq \mu(s_\C, a) = \mathbb{E}[\pi(S, a)\,|\,S_\C = s_\C] = \sum_{s' \in \S^+} p^{\pi}(s' \mid s_\C)\pi(s', a).
    \end{equation}
\end{definition}

\begin{remark}
    By substituting $\C = \emptyset$ into $\tilde\pi_s^a(\C)$, the probability of choosing action $a$ when \emph{all} features are unknown is given by:
    \begin{equation}\nonumber
        \tilde\pi_s^a(\emptyset) = \mathbb{E}[\pi(S, a)].
    \end{equation}
    The expectation is taken over the marginal steady-state distribution $p^\pi(S)$: the probability that an agent following policy $\pi$ is in state $s$.
\end{remark}

When only the features in $\C$ are known, the policy $\mu$ provides the closest approximation to the original policy $\pi$. As such, the difference between $\mu(s_\C, a)$ and $\pi(s, a)$ quantifies how necessary the unknown features $\bar\C$ are to reproduce the behaviour of policy $\pi$; this reflects the influence of those unknown features. Defining the characteristic function as $\tilde\pi_s^a(\C) = \mu(s_\C, a)$ thus ensures that the differences across feature subsets used to compute Shapley values reflect the influence of the unknown features, making it a principled choice for the discrete policy game. Because this construction relies only on the agent’s policy, not on how it was derived, it is general to any policy---optimal and non-optimal.

Unlike defining new learning problems with missing features, or imputing them with a masking value, representing unknown features by marginalising over their conditional distribution does not require changing the MDP or training process. An equally general but alternative approach to representing unknown feature values is to marginalise over their joint distribution:
\begin{equation}
    \tilde\pi_s^a(\C) = \mathbb{E}[\pi(s_\C, S_{\bar\C}, a)] = \sum_{s' \in \S^+} p^\pi(s') \pi(\tau(s, s', \C), a),
\end{equation}
where $\tau(s, s', \C)_i = s_i$ if $i \in \C$, and $s'_i$ otherwise. Unknown feature values are replaced with values sampled from the joint marginal distribution over states.

This approach differs from marginalising over the conditional steady-state distribution in much the same way as in supervised learning (\cref{subsec:sv_sl}). It assumes feature independence and yields equivalent results when that assumption holds. As a result, it assigns zero Shapley value to any feature not explicitly used by the policy. This outcome may appear desirable, but in most function approximation settings, it is less meaningful: such models tend to incorporate all available features to some degree, making true feature irrelevance rare except in contrived examples. When features are dependent, joint marginalisation may produce combinations of values that do not correspond to any valid state. Shapley value attribution relies on measuring behavioural changes in response to changes in known feature values; however, querying a policy with an undefined state renders these changes uninterpretable. For this reason, we focus on the conditional approach in this work, although adapting our framework for joint marginalisation is straightforward. The choice of characteristic function may ultimately depend on the user, but to interpret the resulting Shapley values correctly, it is of paramount importance that they understand the underlying assumptions.

\subsection{An Illustrative Example}
\label{subsec:discretepolicyexample}

We return to the example in \cref{sec:erl,fig:roadsigns}, where an autonomous vehicle navigates junctions using road signs that indicate the direction and distance to the destination. These two values---direction and distance---are the state features, and we represent each state as a tuple $(\text{direction}, \text{distance})$. State 1 has an arrow pointing right and a distance of 10 miles; it is denoted $(\text{R}, 10)$. {State 2} is reached by turning right at state 1; it has an arrow pointing left, and a distance of 10 miles; it is denoted $(\text{L}, 2)$. 

We explain the agent’s behaviour in these two states by representing behaviour under unknown features by marginalising over the conditional steady-state distribution, as defined in \cref{subsec:policycharacteristic}. To do so, we first compute the steady-state distribution of the policy. Starting in state 1, the agent turns right, then left at state 2, and terminates thereafter. This yields a steady-state distribution probability of $0.5$ for each of the two states. Using this distribution, we compute the agent’s behaviour under all subsets of observed features and use the resulting characteristic values to calculate Shapley values for each action, shown in \cref{tab:conditional_example}. They reveal that direction and distance contribute equally to the action taken in both states. This result reflects the behaviour of marginalising over the conditional steady-state distribution: for this policy, either distance or direction is sufficient to identify the action taken, so they are treated as similarly informative and assigned symmetric credit.

\begin{table}[h]
\centering
\renewcommand{\arraystretch}{1.15}
\begin{tabular}{cc|c|cccc|cc}
\toprule
\textbf{State} & \textbf{Action} & $p^\pi(s)$ & $\tilde\pi_s^a(\text{Both})$ & $\tilde\pi_s^a(\text{Dir})$ & $\tilde\pi_s^a(\text{Dist})$ & $\tilde\pi_s^a(\emptyset)$ & $\phi_{\text{Dir}}$ & $\phi_{\text{Dist}}$ \\
\midrule
\multirow{2}{*}{$(\text{R}, 10)$} & L & \multirow{2}{*}{$0.5$} & $0$ & $0$ & $0$ & $0.5$ & $-0.25$ & $-0.25$ \\
                 & R &  & $1$ & $1$ & $1$ & $0.5$ & $\phantom{-}0.25$ & $\phantom{-}0.25$ \\
\midrule
\multirow{2}{*}{$(\text{L}, 2)$} & L & \multirow{2}{*}{$0.5$} & $1$ & $1$ & $1$ & $0.5$ & $\phantom{-}0.25$ & $\phantom{-}0.25$ \\
                & R &  & $0$ & $0$ & $0$ & $0.5$ & $-0.25$ & $-0.25$ \\

\bottomrule
\end{tabular}
\caption{Characteristic values and Shapley values for each state-action pair, computed by marginalising over the conditional steady-state distribution. 
}
\label{tab:conditional_example}
\end{table}

\subsection{Explaining Behaviour in Continuous Action Spaces}
\label{subsec:continuousactions}

So far, we have considered agent behaviour in terms of how feature values influence the probabilities of selecting discrete actions. In continuous action spaces, however, the probability of selecting any particular action is generally zero. As such, policies cannot be interpreted as assigning probabilities to individual actions, and the approach developed in the previous subsections does not directly apply. Instead, we consider an alternative perspective tailored to continuous action settings.

A common policy representation in continuous control is a fixed-variance Gaussian, where actions are sampled from a distribution $\mathcal{N}(\mu(s), \sigma^2)$, where $\mu: \S \rightarrow \mathbb{R}$ maps each state to the mean of the distribution. In this setting, a sound approach to explaining agent behaviour is to analyse how feature values determine the policy's expected action $\mu(s)$. For example, how does a pendulum’s angular velocity influence the expected torque applied in a pendulum balancing task?

This approach naturally leads to the question: how much does each individual feature contribute to the expected action $\mu(s)$? As before, we take a game-theoretic perspective, viewing $\mu(s)$ as the outcome of a coalitional game played by the state features, with Shapley values as the unique solution that satisfies four axioms defining a fair credit allocation.

\begin{definition}[Continuous Policy Game]
    A set $\F = \{1, \dots, n\}$ of features and a characteristic function $\tilde\mu_s: 2^{|\F|} \rightarrow \mathbb{R}$, where $\tilde\mu_s(\C)$ returns the expected action in state $s$ when only the features in $\C \subseteq \F$ are known.
\end{definition}

Central to the continuous policy game $(\F, \tilde\mu_s)$ is the characteristic function $\tilde\mu_s$, which extends the mean function $\mu$ to situations where only a subset of state features is known. The precise construction of $\tilde\mu_s$ will be addressed later in this section; for now, we assume it is given and turn to our central question: what is the contribution of each feature value $s_i \in \S_i$ to the policy's expected action $\mu(s)$? We denote this contribution by $\phi_i(\tilde\mu_s)$. In other words, how can we fairly allocate---across the state features---the difference between the expected action when all features are known and when none are? We again appeal to Shapley's foundational axioms, restated here to clarify their role in this specific attribution setting. Below, we define four axioms that any attribution $\phi_i(\tilde\mu_s)$, for all $i \in \mathcal{F}$, should satisfy to qualify as a fair credit allocation.

\begin{axiom}[Efficiency]
    The difference in value between the expected action when all features are known, $\tilde\mu_s(\F)$, and the expected action when no features are known, $\tilde\mu_s(\emptyset)$, is fully distributed across the features:
    \begin{equation}\nonumber
        \tilde\mu_s(\F) - \tilde\mu_s(\emptyset) = \sum_{i \in \F} \phi_i(\tilde\mu_s).
    \end{equation}
\end{axiom}

\begin{axiom}[Linearity]
    For a composite policy that samples actions from a Gaussian whose mean $\alpha\,\tilde\mu(s) + (1 - \alpha)\,\tilde\nu(s)$ is the weighted average of the mean functions $\tilde\mu(s)$ and $\tilde\nu(s)$ for two separate policies, a feature's contribution to the combined policy's expected action is equal to the weighted sum of its contributions to each individual policy's expected action:
    \begin{equation}\nonumber
        \phi_i(\alpha\,\tilde\mu_s + (1 - \alpha)\,\tilde\nu_s) = \alpha\,\phi_i(\tilde\mu_s) + (1 - \alpha)\,\phi_i(\tilde\nu_s).
    \end{equation}
\end{axiom}

\begin{axiom}[Symmetry]
    Two features are assigned the same contribution to the expected action if they make equal marginal contributions for all possible coalitions:
    \begin{equation}\nonumber
        \phi_i(\tilde\mu_s) = \phi_j(\tilde\mu_s) \quad \text{if} \quad \tilde\mu_s(\C \cup \{i\}) = \tilde\mu_s(\C \cup \{j\}) \quad \forall \C \subseteq \F \setminus \{i, j\}.
    \end{equation}
\end{axiom}

\begin{axiom}[Nullity]
    A feature is assigned zero contribution if it makes zero marginal contribution to the expected action for all possible coalitions:
    \begin{equation}\nonumber
        \phi_i(\tilde\mu_s) = 0 \quad \text{if} \quad \tilde\mu_s(\C \cup \{i\}) = \tilde\mu_s(\C) \quad \forall \C \subseteq \F \setminus \{i\}.
    \end{equation}
\end{axiom}

These four axioms are directly analogous to those used in the discrete case and identify the Shapley values as the unique solution for attributing the mean $\mu(s)$ across features.

\begin{proposition}\label{prop:continuouspolicyshapley}
    The Shapley values $\phi_1(\tilde\mu_s), \dots, \phi_n(\tilde\mu_s)$ for the continuous policy game $(\F, \tilde\mu_s)$ are the unique allocation of the expected action $\mu(s)$ across the features of state $s$ that satisfies the four axioms of fair credit assignment: efficiency, linearity, symmetry, and nullity.
\end{proposition}

The Shapley values of the continuous policy game provide a principled way to attribute an agent's expected action to individual state features, offering a concrete explanation of behaviour in continuous action spaces. As in the discrete case, computing Shapley values requires defining the characteristic function $\tilde\mu_s(\C)$, which entails making assumptions about how the agent behaves when only partial state information is available.

\begin{definition}
    Let $\zeta: \S_\C \rightarrow \Delta(\A)$ be a policy defined over a subset of features $\C$, where actions are sampled from a fixed-variance Gaussian $\mathcal{N}(\omega(s_\C), \sigma^2)$. The mean function $\omega: \S_\C \rightarrow \mathbb{R}$ defines the expected action when only the features in $\C$ are known:
    \begin{equation}\nonumber
        \omega(s_\C) = \mathbb{E}[\mu(S)\,|\,S_\C = s_\C] = \sum_{s' \in \S^+} p^\pi(s' \mid s_\C)\,\mu(s').
    \end{equation}
\end{definition}

Among all fixed-variance Gaussian policies defined over $\C$, the policy $\zeta$ is the one whose mean $\omega$ deviates the least from $\mu$ in expectation.

\begin{proposition}\label{prop:continuouspolicychar}
    When comparing mean functions using mean squared error, the function $\omega$ deviates the least in expectation from the mean $\mu$ of the full policy $\pi$:
    \begin{equation}\label{eq:omega}
        \omega = \argmin_{g: \S_\C \rightarrow \mathbb{R}} \mathbb{E}\left[\lVert \mu(S) - g(S_\C) \rVert^2\right].
    \end{equation}
\end{proposition}

This motivates our definition of the characteristic function $\tilde\mu_s(\C)$: the expected action when only the features in $\C$ are known is taken to be the value given by $\omega(s_\C)$ at that state.

\begin{definition}[Continuous Policy Characteristic Function]
    The expected action when only the features in $\C$ are known is defined as:
    \begin{equation}\label{eq:policychar}
        \tilde\mu_s(\C) \defeq \omega(s_\C) = \mathbb{E}[\mu(S) \mid S_\C = s_\C] = \sum_{s' \in \S^+} p^\pi(s' \mid s_\C)\,\mu(s').
    \end{equation}
\end{definition}

When only the features in $\C$ are known, the function $\tilde\mu_s(\C)$ provides the closest approximation to the full policy's expected action $\mu(s)$. As such, the difference between $\tilde\mu_s(\C)$ and $\mu(s)$ quantifies how necessary the unknown features $\bar\C$ are to reproduce the agent’s behaviour. Defining the characteristic function this way ensures that the differences across feature subsets used to compute Shapley values reflect the influence of the unknown features, making it a principled choice for the continuous policy game. This construction is general to any policy.

As in the discrete setting, alternative characteristic functions may also be considered here. In particular, a marginal variant of $\tilde\mu_s(\C)$---which assumes feature independence---could be used in place of the conditional formulation discussed above.

We do not present an illustrative example in this setting, as doing so would require estimating the steady-state distribution and conditional expectations in continuous state-spaces---a task beyond this work's scope. Nonetheless, the theory developed here offers a general and principled foundation for future approaches to explaining agent behaviour in continuous action domains.

%% file: sections/5_explaining_outcomes.tex
\section{Explaining Outcomes}
\label{sec:explainingoutcome}

Outcomes can refer to many measurable consequences of an agent’s behaviour, such as next states or immediate rewards. Here, we focus on expected return, a standard formalisation of outcome in reinforcement learning. Other forms of outcomes may be addressed in future work. The outcome of a policy $\pi$ can be quantified by the state-value function $v^\pi: \S \rightarrow \mathbb{R}$, which gives the expected return from each state when following that policy:
\begin{equation}
    v^\pi(s) \defeq \mathbb{E}_\pi\left[G_t \mid S_t = s\right].
\end{equation}
Given that states are composed of feature values, it is natural to ask how individual features contribute to this expected return. For example, how does a navigation agent’s $x-$coordinate influence the expected return of its policy at a given location? In this section, we develop a theoretical framework for attributing expected return to the values of state features, offering a structured explanation of agent outcomes under policy~$\pi$.

\subsection{State Features Collaborate to Determine Expected Return}
\label{subsec:outcomeshapley}

As with explaining agent behaviour, we start by taking a game-theoretic view of expected return. Just as an agent’s policy arises from the combined influence of its state features, the expected return $v^\pi(s)$ can also be viewed as the outcome of a collaborative process among those features. That is, the return achieved from a state is affected by how the policy responds to the information available at that state. This perspective motivates the question: how much does each individual feature contribute to the overall return? To answer this, we define another coalitional game over state features, introduce a new set of fairness axioms, and show that Shapley values offer a unique principled solution.

\begin{definition}[Outcome Game]
    A set $\F = \{1, \dots, n\}$ of state features and a characteristic function $\tilde{v}^\pi_s: 2^{|\F|} \rightarrow \mathbb{R}$, where $\tilde{v}^\pi_s(\C)$ denotes the expected return from state $s$ when policy $\pi$ has access to only the values of features in $\C \subseteq \F$.
\end{definition}

The characteristic function $\tilde{v}^\pi_s$ extends the value function $v^\pi$ to settings where only a subset of feature values are accessible. As before, we postpone the precise construction of $\tilde{v}^\pi_s(\C)$ until later. For now, we ask: what is the contribution of each feature value $s_i \in \S_i$ to the expected return $v^\pi(s)$? We denote this by $\phi_i(\tilde{v}^\pi_s)$. We appeal to Shapley's foundational axioms, restated here to clarify their role in this specific attribution setting, and identify the Shapley values as the unique solution.

\begin{axiom}[Efficiency]
The difference in expected return between the policy with all features known and the policy with none known is fully distributed across the features:
\begin{equation}\nonumber
    \tilde{v}^\pi_s(\F) - \tilde{v}^\pi_s(\emptyset) = \sum_{i \in \F} \phi_i(\tilde{v}^\pi_s).
\end{equation}
\end{axiom}

\begin{axiom}[Linearity]
For two independent policies $\pi$ and $\omega$, a feature’s contribution to the weighted sum of their expected returns is equal to the corresponding weighted sum of the feature's contributions under each policy:
\begin{equation}\nonumber
    \phi_i(\alpha\,\tilde{v}^\pi_s + \beta\,\tilde{v}^\omega_s) = \alpha\,\phi_i(\tilde{v}^\pi_s) + \beta\,\phi_i(\tilde{v}^\omega_s).
\end{equation}
\end{axiom}

\begin{axiom}[Symmetry]
Two features are assigned the same contribution if they make equal marginal contributions to the expected return for all possible coalitions:
\begin{equation}\nonumber
    \phi_i(\tilde{v}^\pi_s) = \phi_j(\tilde{v}^\pi_s) \quad \text{if} \quad \tilde{v}^\pi_s(\C \cup \{i\}) = \tilde{v}^\pi_s(\C \cup \{j\}) \quad \forall \C \subseteq \F \setminus \{i, j\}.
\end{equation}
\end{axiom}

\begin{axiom}[Nullity]
A feature is assigned zero contribution if it makes no marginal contribution to the expected return for all possible coalitions:
\begin{equation}\nonumber
    \phi_i(\tilde{v}^\pi_s) = 0 \quad \text{if} \quad \tilde{v}^\pi_s(\C \cup \{i\}) = \tilde{v}^\pi_s(\C) \quad \forall \C \subseteq \F \setminus \{i\}.
\end{equation}
\end{axiom}

These four axioms mirror those used to explain behaviour and are directly analogous to the standard axioms from cooperative game theory.

\begin{proposition}\label{prop:outcomeshapley}
    The Shapley values $\phi_1(\tilde{v}^\pi_s), \dots, \phi_n(\tilde{v}^\pi_s)$ for the outcome game $(\F, \tilde{v}^\pi_s)$ are the unique allocation of the expected return $v^\pi(s)$ across the features of state~$s$ that satisfies the four axioms of fair credit assignment: efficiency, linearity, symmetry, and nullity.
\end{proposition}

The Shapley values of the outcome game provide a principled way to attribute expected return to individual state features, offering a concrete explanation of agent outcomes at a given state.

We now revisit the example from \cref{sec:erl,fig:roadsigns}, where an autonomous vehicle navigates junctions using road signs that indicate the direction and distance to a destination. Each state is represented as a tuple $(\text{direction}, \text{distance})$. The agent's policy moves right at the first state: $(\text{R}, 10)$, and left at the next state: $(\text{L}, 2)$. We use an undiscounted reward function of $-1$ for each action, with an additional reward of $+10$ for reaching the goal---a shortest-path problem. An episode terminates if the agent reaches the goal or turns off the map (e.g. turning right at the second state).

We use the proposed framework to attribute the expected return from each state to its observed feature values. However, doing so requires an assumption about how unknown features affect expected return. To motivate this assumption, note that any change in expected return due to missing features must arise from a shift in the agent's behaviour at that state, assuming behaviour and dynamics at all other states remain fixed. Based on this reasoning, we assume that the expected return with unknown features corresponds to the expected return from acting with partial information in the given state and full observability elsewhere. To specify how the agent acts with partial information, we reuse the two agents introduced earlier: \texttt{Agent A} follows the direction if visible and acts randomly otherwise. \texttt{Agent B} uses both direction and distance, turning right if it observes 10 miles or a right direction, left if it observes 2 miles or a left direction, and moving randomly otherwise. These assumptions define the policy under partial information and thus determine the expected return from each state for every feature subset. The resulting characteristic and Shapley values are shown in \cref{tab:outcome_example_A,tab:outcome_example_B}.

\begin{table}[h]
\centering
\renewcommand{\arraystretch}{1.15}
\begin{tabular}{c|cccc|cc}
\toprule
\textbf{State $(s)$} & $\tilde{v}^\pi_s(\text{Both})$ & $\tilde{v}^\pi_s(\text{Dir})$ & $\tilde{v}^\pi_s(\text{Dist})$ & $\tilde{v}^\pi_s(\emptyset)$ & $\phi_{\text{Dir}}$ & $\phi_{\text{Dist}}$ \\
\midrule
$(\text{R}, 10)$ & $8$ & $8$ & $8$ & $8$ & $0$ & $0$ \\
$(\text{L}, 2)$  & $9$ & $9$ & $4$ & $4$ & $5$ & $0$ \\
\bottomrule
\end{tabular}
\caption{Characteristic values and Shapley values for expected return under \texttt{Agent A}.}
\label{tab:outcome_example_A}
\end{table}

\begin{table}[h]
\centering
\renewcommand{\arraystretch}{1.15}
\begin{tabular}{c|cccc|cc}
\toprule
\textbf{State $(s)$} & $\tilde{v}^\pi_s(\text{Both})$ & $\tilde{v}^\pi_s(\text{Dir})$ & $\tilde{v}^\pi_s(\text{Dist})$ & $\tilde{v}^\pi_s(\emptyset)$ & $\phi_{\text{Dir}}$ & $\phi_{\text{Dist}}$ \\
\midrule
$(\text{R}, 10)$ & $8$ & $8$ & $8$ & $8$ & $0$ & $0$ \\
$(\text{L}, 2)$  & $9$ & $9$ & $9$ & $4$ & $2.5$ & $2.5$ \\
\bottomrule
\end{tabular}
\caption{Characteristic values and Shapley values for expected return under \texttt{Agent B}.}
\label{tab:outcome_example_B}
\end{table}

Consider first the state $(\text{R}, 10)$. For both agents, neither feature contributes to expected return. The reason is that turning left or right from the initial junction results in paths of equal length to the goal, yielding identical returns. In contrast, in our explanation of behaviour, the direction feature received credit for prompting the agent to turn right. This example demonstrates a key distinction between explaining behaviour and outcomes: a feature can influence behaviour without affecting its outcomes.

In the second state, $(\text{L}, 2)$, the features do influence expected return. For \texttt{Agent A}, all credit goes to the direction feature, reflecting its exclusive influence on the agent’s decision. For \texttt{Agent B}, both features contribute equally because either can be used to determine the correct action. These outcome attributions mirror our behaviour analysis under the same assumptions, illustrating how the two forms of explanation can align.


\subsection{What Can Be Assumed About Expected Return When Some State Features Are Unknown?}
\label{subsec:outcomecharacteristic}

We now consider how to define the characteristic function $\tilde{v}^\pi_s(\C)$: the expected return from state $s$ when the agent’s policy has access only to the features in $\C$. Since the value function $v^\pi(s)$ assumes full observability, it does not specify this quantity, so we must make additional assumptions. 

To motivate our definition, recall from the previous example that any change in expected return due to unknown features must reflect a shift in the agent’s behaviour at state $s$, assuming all else remains fixed. We therefore define expected return under unknown features as the expected return when the agent acts with partial information whenever it visits state $s$, while following its fully observed policy in all other states. This ensures that any change in return reflects only how the agent behaves differently at state $s$ under partial information, not how this might influence behaviour elsewhere in the environment. In this sense, the construction isolates the contribution of features at state~$s$. It also enables us to express $\tilde{v}^\pi_s(\C)$ using the discrete policy characteristic function $\tilde\pi_s^a(\C)$ introduced in \cref{subsec:policycharacteristic}.

Recall that $\tilde\pi_s^a(\C)$ defines the probability of selecting action $a$ in state $s$ when only the features in $\C$ are known:
\begin{equation}\nonumber
    \tilde\pi_s^a(\C) = \mathbb{E}[\pi(S, a)\,|\,S_\C = s_\C].
\end{equation}
These probabilities form a complete action distribution at $s$ under partial information, which we can use to evaluate expected return with unknown features.

\begin{definition}[Outcome Characteristic Function]
    The expected return from state $s$ when an agent has access only to features in $\C$ is defined as:
    \begin{equation}\label{eq:outcomechar}
        \tilde{v}^\pi_s(\C) = \mathbb{E}_\mu \left[G_t \mid S_t = s\right],
    \end{equation}
    where the policy $\mu$ is defined as
    \begin{equation}\nonumber
        \mu(s_t, a_t) =
        \begin{cases}
            \pi_{s_t, a_t}(\C) & \text{if } s_t = s, \\
            \pi(s_t, a_t) & \text{otherwise.}
        \end{cases}
    \end{equation}
\end{definition}

This definition ensures that unknown features affect only how the agent behaves whenever it visits state $s$, while it follows its original policy $\pi$ in all other states. As a result, any change in expected return can be attributed specifically to the unknown features at state~$s$, isolating their contribution.

\begin{remark}
    Since we defined $\tilde\pi_s^a(\C)$ as a marginal over the full state space, it may assign nonzero probability to unavailable actions in $s$. In such cases, the distribution over $\A(s)$ should be renormalised with zero probability assigned to the unavailable actions.
\end{remark}

From Proposition \ref{prop:policychar}, $\tilde\pi_s^a(\C)$ is the best approximation to the original policy $\pi$ when only the features in $\C$ are available; the expected return $\tilde{v}^\pi_s(\C)$ captures the outcome of following this approximation under the partial information. The difference between $\tilde{v}^\pi_s(\C)$ and $v^\pi(s)$ thus quantifies how necessary the unknown features $\bar\C$ are for achieving the original expected return. Defining the characteristic function in this way ensures that the differences across feature subsets used to compute Shapley values capture the influence of the unknown features, making it a principled choice for the outcome game. As before, this construction is general to any policy.

Alternative definitions of agent behaviour under partial information could also be substituted into our proposed characteristic function. For example, the marginal variant of $\tilde\pi_s^a(\C)$---which assumes feature independence---may be appropriate in some settings; see \cref{subsec:policycharacteristic} for further discussion. Similarly, in continuous action spaces, the policy with partial information can be defined as a Gaussian with mean $\mu_s(\C)$, from \cref{eq:policychar}, in place of the discrete policy based on $\tilde\pi_s^a(\C)$; this substitution naturally extends the characteristic function $\tilde{v}^\pi_s(\C)$ to continuous domains. We leave a detailed treatment of this extension to future work.

We do not present a new illustrative example for the proposed characteristic function because its outcome is identical to the example in the previous subsection. In the road-sign domain, the discrete policy characteristic $\tilde\pi_s^a(\C)$ behaves exactly like \texttt{Agent B} under partial information. Since the outcome characteristic $\tilde{v}^\pi_s(\C)$ is defined using this same policy characteristic, it leads to the same values and attributions already shown for \texttt{Agent B} in \cref{tab:outcome_example_B}. As such, the example already demonstrates how Shapley values enable intuitive explanations of agent outcomes using this characteristic function. We provide an additional example illustrating how SVERL explains agent outcomes in \cref{sec:additionalexamples}.

%% file: sections/6_explaining_predictions.tex
\section{Explaining Predictions} 
\label{sec:explainingvalueestimation}


Just as outcomes can refer to different consequences of agent behaviour, such as next states or rewards, predictions can target different estimates of these same quantities. In this section, we focus on predicting expected return, in line with our treatment of outcomes. Expected return can be formalised using either the state-value function $v^\pi(s)$ or the state-action value function $q^\pi(s, a)$; here, we focus on the former. Specifically, we consider how an agent or observer predicts $v^\pi(s)$, the expected return under policy $\pi$ from a given state. As before, we aim to understand how individual feature values contribute to this estimate. Analogous explanations for $q^\pi(s, a)$ and other prediction targets can be developed similarly.

At first glance, predictions may seem indistinguishable from explaining outcomes: both concern expected return under a given policy. However, the distinction lies in what role the feature values play. When explaining outcomes, we assess how features influence expected return through their effect on the agent’s behaviour. In contrast, when explaining predictions, we assess how features influence an external estimate of expected return, regardless of whether those features affect the agent’s behaviour.

To formalise this distinction, we introduce $\hat{v}^\pi: \S \rightarrow \mathbb{R}$, denoting an estimate of $v^\pi(s)$ made by the agent or an observer. When the estimate is accurate, $\hat{v}^\pi(s) = v^\pi(s)$, but our goal is not to evaluate its accuracy. Rather, by treating $\hat{v}^\pi$ as a predictive function, we isolate the epistemic question: how do feature values contribute to what we predict an agent to achieve, without assuming they shape what the agent actually does?

\subsection{State Features Collaborate to Determine Value Estimates}
\label{subsec:valuepredshapley}

We now take a game-theoretic view of predictions. The predicted value $\hat{v}^\pi(s)$ can be viewed as the outcome of a collaborative process among features. That is, each feature value in $s$ contributes to the prediction of expected return under policy~$\pi$. To attribute this prediction across the feature values, we define another coalitional game over the features, introduce a new set of fairness axioms tailored to the prediction setting, and identify the Shapley values as the unique solution.

\begin{definition}[Prediction Game]
    A set $\F = \{1, \dots, n\}$ of features and a characteristic function $\hat{v}^\pi_s: 2^{|\F|} \rightarrow \mathbb{R}$, where $\hat{v}^\pi_s(\C)$ denotes the predicted expected return using only the feature values $s_\C \in \S_\C$.
\end{definition}

As before, we postpone the precise construction of the characteristic function $\hat{v}^\pi_s$ until later. For now, we ask: what is the contribution of each feature value $s_i \in \S_i$ to the estimate $\hat{v}^\pi(s)$? We denote this contribution by $\phi_i(\hat{v}^\pi_s)$ and appeal to Shapley’s foundational axioms, restated here to clarify their role in this specific attribution setting.

\begin{axiom}[Efficiency]
The difference in value between the predicted expected return using all features, $\hat{v}^\pi_s(\F)$, and when using none, $\hat{v}^\pi_s(\emptyset)$, is fully distributed across the features:
\begin{equation}\nonumber
    \hat{v}^\pi_s(\F) - \hat{v}^\pi_s(\emptyset) = \sum_{i \in \F} \phi_i(\hat{v}^\pi_s).
\end{equation}
\end{axiom}

\begin{axiom}[Linearity]
For two independent policies $\pi$ and $\omega$, a feature’s contribution to the weighted sum of their predicted expected returns is equal to the corresponding weighted sum of the feature's contributions under each policy:
\begin{equation}\nonumber
    \phi_i(\alpha\,\hat{v}^\pi_s + \beta\,\hat{v}^\omega_s) = \alpha\,\phi_i(\hat{v}^\pi_s) +  \beta\,\phi_i(\hat{v}^\omega_s).
\end{equation}
\end{axiom}

\begin{axiom}[Symmetry]
Two features receive equal attribution if they make identical marginal contributions to the predicted expected return for all possible coalitions:
\begin{equation}\nonumber
    \phi_i(\hat{v}^\pi_s) = \phi_j(\hat{v}^\pi_s) \quad \text{if} \quad \hat{v}^\pi_s(\C \cup \{i\}) = \hat{v}^\pi_s(\C \cup \{j\}) \quad \forall \C \subseteq \F \setminus \{i, j\}.
\end{equation}
\end{axiom}

\begin{axiom}[Nullity]
A feature receives zero attribution if it makes no marginal contribution to the predicted expected return in any coalition:
\begin{equation}\nonumber
    \phi_i(\hat{v}^\pi_s) = 0 \quad \text{if} \quad \hat{v}^\pi_s(\C \cup \{i\}) = \hat{v}^\pi_s(\C) \quad \forall \C \subseteq \F \setminus \{i\}.
\end{equation}
\end{axiom}

These four axioms mirror those used when explaining behaviour and outcomes.

\begin{proposition}\label{prop:valueshapley}
    The Shapley values $\phi_1(\hat{v}^\pi_s), \dots, \phi_n(\hat{v}^\pi_s)$ for the prediction game $(\F, \hat{v}^\pi_s)$ are the unique allocation of the predicted expected return $\hat{v}^\pi(s)$ across the features of state~$s$ that satisfies the four axioms of fair credit assignment: efficiency, linearity, symmetry, and nullity.
\end{proposition}

The Shapley values of the prediction game provide a principled way to attribute state-value estimates to individual state features, offering a structured explanation of predictions. As with the previous coalitional games, computing these Shapley values requires precisely defining the characteristic function $\hat{v}^\pi_s(\C)$, which we turn to next.

\subsection{What Can Be Assumed About Value Estimates When Some State Features Are Unknown?}
\label{subsec:valueestimationcharacteristic}

We now consider how to define the characteristic function $\hat{v}^\pi_s(\C)$: the predicted expected return from state $s$ when using only the features in $\C \subseteq \F$. A natural principle is to best estimate value using only the information that is actually available. The predicted expected return should reflect the best assessment based on the observed features alone.

\begin{definition}
    Let $\hat{u}^\pi: \S_\C \rightarrow \mathbb{R}$ be a value function defined over a subset of features $\C \subseteq \F$ such that the estimated expected return from state $s$ when only the features in $\C$ are observed is given by:
    \begin{equation}\nonumber
        \hat{u}^\pi(s_\C) = \mathbb{E}[\hat{v}^\pi(S) \mid S_\C = s_\C] = \sum_{s' \in \S^+} p^\pi(s' \mid s_\C)\,\hat{v}^\pi(s').
    \end{equation}
\end{definition}

Among all functions defined over the observed features, $\hat{u}^\pi$ is the one that deviates the least, in expectation, from the full-information estimate $\hat{v}^\pi$.

\begin{proposition}\label{prop:valuechar}
    When comparing expected-return estimates using mean squared error, the function $\hat{u}^\pi$ is the closest in expectation to $\hat{v}^\pi$ among all functions defined over the subset of features $\C$:
    \begin{equation}\nonumber
        \hat{u}^\pi = \argmin_{g: \S_\C \rightarrow \mathbb{R}} \mathbb{E}\left[\lVert \hat{v}^\pi(S) - g(S_\C) \rVert^2\right].
    \end{equation}
\end{proposition}

This motivates our definition of the characteristic function for predictions:

\begin{definition}[Prediction Characteristic Function]
    The predicted expected return from state $s$ when using only the features in $\C$ is given by:
    \begin{equation}\label{eq:valuechar}
        \hat{v}^\pi_s(\C) \defeq \hat{u}^\pi(s_\C) = \mathbb{E}[\hat{v}^\pi(S) \mid S_\C = s_\C] = \sum_{s' \in \S^+} p^\pi(s' \mid s_\C)\,\hat{v}^\pi(s').
    \end{equation}
\end{definition}

When only the features in $\C$ are observed, the function $\hat{u}^\pi$ provides the best approximation of the full-information value estimate $\hat{v}^\pi(s)$. As such, the difference between $\hat{u}^\pi(s_\C)$ and $\hat{v}^\pi(s)$ quantifies how necessary the unobserved features $\bar\C$ are for recovering the original estimate. Defining the characteristic function as $\hat{v}^\pi_s(\C) = \hat{u}^\pi(s_\C)$ ensures that the differences across feature subsets used to compute Shapley values reflect the influence of the unknown features, making it a principled choice for the prediction game.

This construction is general to any policy. Alternative characteristic functions could also be used. In particular, a marginal variant of $\hat{v}^\pi_s(\C)$---which assumes feature independence---may be suitable in some settings. An analogous discussion of this alternative appears for the discrete policy characteristic function in \cref{subsec:policycharacteristic}.

\subsection{Illustrative Example}
\label{subsec:valueestimationexample}

We return to the earlier navigation example (\cref{sec:erl,fig:roadsigns}) with features $direction$ and $distance$. 
Using the prediction characteristic function, we apply the proposed framework to analyse how value can be predicted in the two states examined earlier, $(\text{R}, 10)$ and $(\text{L}, 2)$. \cref{tab:valueestimation_example} presents the predicted expected return for different feature subsets, and the resulting Shapley values for each feature.

\begin{table}[h]
\centering
\renewcommand{\arraystretch}{1.15}
\begin{tabular}{c|c|cccc|cc}
\toprule
\textbf{State ($s$)} & $p^\pi(s)$ & $\hat{v}^\pi_s(\text{Both})$ & $\hat{v}^\pi_s(\text{Dir})$ & $\hat{v}^\pi_s(\text{Dist})$ & $\hat{v}^\pi_s(\emptyset)$ & $\phi_{\text{Dir}}$ & $\phi_{\text{Dist}}$ \\
\midrule
$(\text{R}, 10)$ & $0.5$ & $8$ & $8$ & $8$ & $8.5$ & $-0.25$ & $-0.25$ \\
$(\text{L}, 2)$  & $0.5$ & $9$ & $9$ & $9$ & $8.5$ & $\phantom{-}0.25$ & $\phantom{-}0.25$ \\
\bottomrule
\end{tabular}
\caption{The steady-state distribution, characteristic values, and Shapley values when explaining predictions for each state in the road sign example of \cref{fig:roadsigns}.}
\label{tab:valueestimation_example}
\end{table}

We first explore the \emph{distance} feature. In state $(\text{R}, 10)$, the agent is 10 miles from the goal. The distance feature reduces the return estimate relative to the baseline with no features, resulting in a negative Shapley value ($-0.25$). In state $(\text{L}, 2)$, the agent is much closer to the goal, and the distance feature increases the return estimate, producing a positive Shapley value ($+0.25$). These attributions reflect an intuitive interpretation: when the agent is far from the goal, the expected return is lower; when it is near, the expected return is higher. The Shapley values capture how each feature contributes to the estimate, appropriately assigning responsibility to the distance feature in line with how it indicates distance to the goal. The \emph{direction} feature receives identical attributions because it also helps differentiate the two visited states. In this domain, the sign’s direction indirectly conveys distance: the policy encounters the right-facing sign only when far from the goal and the left-facing sign only when close.


\subsection{Contrasting Predictions with Outcomes}
\label{subsec:contrast}

\begin{figure}[t]
    \centering
    \includegraphics[width=0.5\columnwidth]{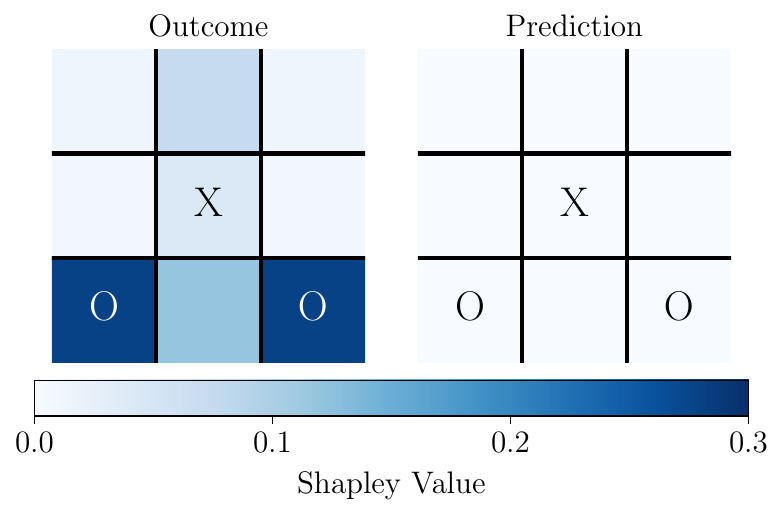}
    \caption{\texttt{Tic-Tac-Toe}. Left: Contributions to the agent's expected return (outcome). Right: Contributions to the agent's estimated expected return (prediction). 
    }
    \label{fig:tic_tac_toe_perf_value}
\end{figure}

With the final component of SVERL introduced, we now ground the distinction between explaining outcomes and predictions in a concrete example using the game of \texttt{Tic-Tac-Toe}. Consider an agent playing optimally against a minimax opponent~\citep{Polak1989}. Against such an opponent, the agent cannot win but can avoid losing. In other words, every game ends in a draw. As a result, the expected return is zero from every state. Now consider the state shown in \cref{fig:tic_tac_toe_perf_value}. The opponent (\texttt{O}) moved first, and it is now the agent's (\texttt{X}) turn. Unless the agent marks the bottom-centre square, it will lose. We consider how each of the nine grid squares contributes to outcomes and predictions.

The left side of \cref{fig:tic_tac_toe_perf_value} shows the SVERL attributions for explaining outcomes---how each of the nine grid squares, taking values ``X'', ``O'', or ``empty'', contributes to the agent’s expected return through their influence on the agent’s policy. The opponent's two squares have the highest attributions, reflecting the agent's need to block the opponent's winning move. The right side of \cref{fig:tic_tac_toe_perf_value} shows the corresponding SVERL attributions for predicting the agent's expected return. All feature contributions are zero. While initially surprising, this is correct: every state leads to a draw, so the expected return is always zero, and thus no individual feature is more or less predictive of the expected return than any other. 

Explaining predictions can reveal features useful for predicting expected return, but it does not explain how feature values shape the agent’s behaviour or the outcomes of that behaviour. In contrast, explaining outcomes reveals how feature values impact expected return as a direct consequence of their impact on policy.

%% file: sections/7_examples.tex
\section{Understanding SVERL Through Examples}
\label{sec:examples}

We present two examples to illustrate how SVERL captures the changing role of information in shaping an agent’s actions, expected returns, and predictions as episodes progress. Additional examples are provided in \cref{sec:additionalexamples}. Unless stated otherwise, all examples are undiscounted and explain optimal policies computed via value iteration \citep{Sutton2018}. Steady-state distributions are either computed exactly or accurately estimated from trajectories; expected returns for the outcome characteristic are computed using value iteration.%
\footnote{Code is available at \url{https://github.com/djeb20/sverl}.}

\subsection{Mastermind}
\label{subsec:mastermind}

\begin{figure}[t]
\begin{center}
\centerline{\includegraphics[width=0.8\columnwidth]{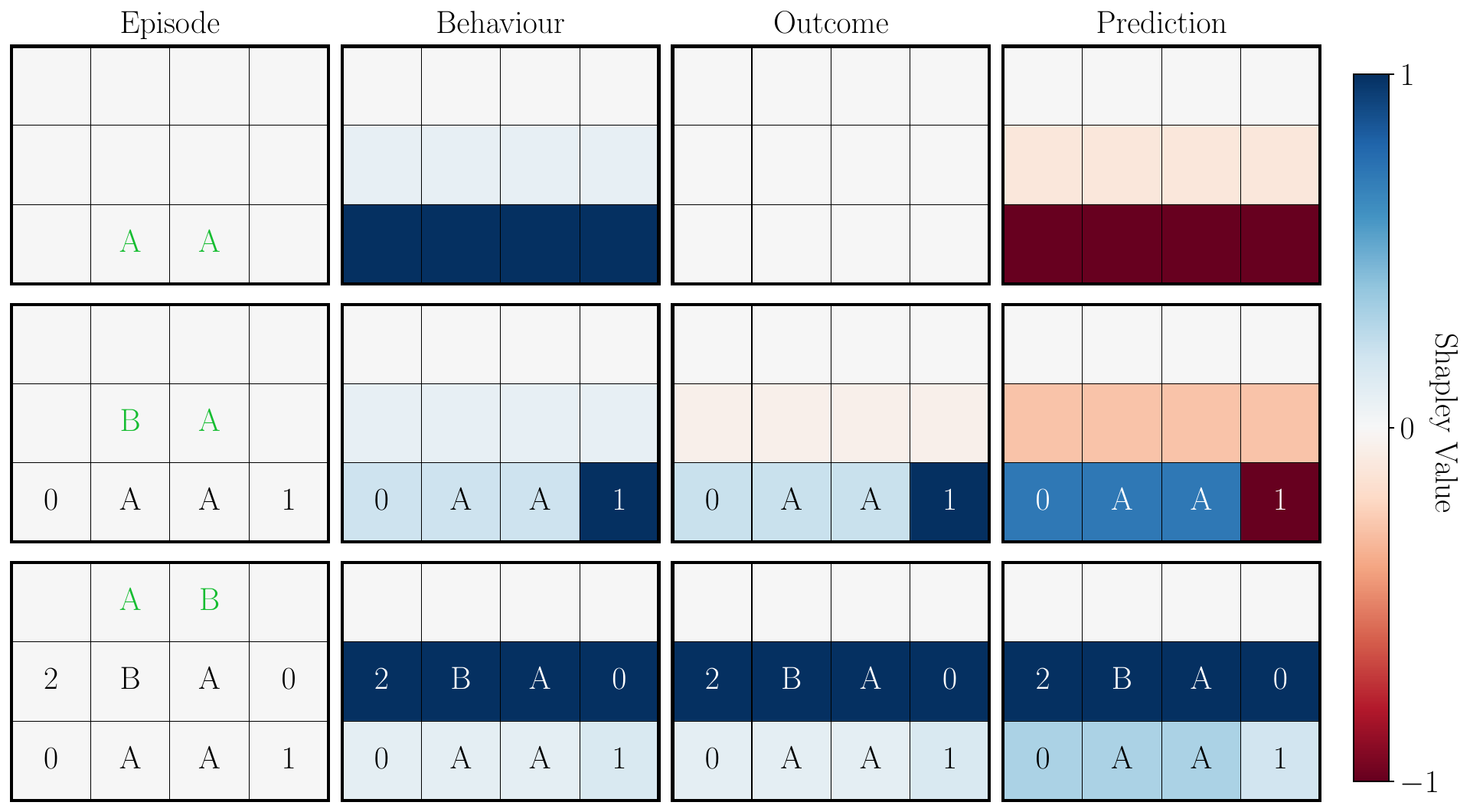}}
\caption{Shapley value attributions for one complete episode of \texttt{Mastermind}, consisting of three states (top to bottom). First column: episode progression, with the optimal action taken in each state shown in green ink---these are not part of the state. Second column: feature contributions to the probability of selecting the optimal action taken (behaviour). Third column: feature contributions to the expected return (outcome). Fourth column: feature contributions to the estimated expected return (prediction). The right-most column on the board displays the position clue; the left-most column displays the misplaced clue. Shapley values are scaled to lie between $-1$ and $1$.}
\label{fig:mastermind_sverl}
\end{center}
\end{figure}

In the classic code-breaking game \texttt{Mastermind}, a secret code is selected, and the player attempts to guess it within a limited number of attempts. In our formulation, each code consists of two letters, where each letter can be either $\mathrm{A}$ or $\mathrm{B}$, resulting in four possible secret codes: $\mathrm{AA}$, $\mathrm{AB}$, $\mathrm{BA}$, and $\mathrm{BB}$. After each guess, the player receives feedback in the form of two clues:
\begin{itemize}
\item[] \textbf{Position clue:} the number of letters in the player's guess that are present in the secret code and are in the correct position.
\item[] \textbf{Misplaced clue:} the number of letters in the player's guess that are present in the secret code but are not in the correct position.
\end{itemize}
The position clue is computed first, and matched letters are excluded when calculating the misplaced clue. For example, if the secret code is $\mathrm{AB}$ and the guess is $\mathrm{AA}$, the position clue is 1 and the misplaced clue is 0.

We formalise this game as a Markov decision process. At the start of each episode, the environment randomly selects one of the four possible codes as the hidden target. The agent interacts with the environment over three decision stages, submitting one guess per stage. After each guess, the environment returns the position and misplaced clues as observation feedback. The episode terminates when the agent guesses the correct code or after three unsuccessful attempts. Incorrect guesses receive a reward of $-1$, and correct guesses receive $0$. Each state is represented by the visible game board, encoded as 16 features consisting of the agent's previous guesses and their associated feedback. If fewer than three guesses have been made, the remaining feature values are set to a special \texttt{empty} token to indicate the absence of information.

We use SVERL to analyse an agent in \texttt{Mastermind} to show how explanations of behaviour, outcomes, and predictions can align or diverge to reveal complementary insights. \cref{fig:mastermind_sverl} shows the state-trajectory for an optimal deterministic policy (left column) and corresponding explanations for an episode where the hidden code is $\mathrm{AB}$.

At the first decision stage, the episode begins on the empty game board. Based on this state, the agent selects $\mathrm{AA}$---though any initial action is optimal. The explanation of this behaviour assigns large positive attributions to the empty bottom row, corresponding to where the agent will place its first code. The prediction explanation assigns large negative attributions to the same features, while the outcome explanation assigns no responsibility to any feature. These patterns reflect that the initial state, indicated by the empty bottom row, has a low value estimate and determines the agent's initial action. They also reflect that the return is invariant to the known features because all possible initial actions yield the same expected return.

At the second decision stage, the agent can conclude that one letter in its initial guess is correctly placed and the other should be a $\mathrm{B}$. Consequently, it selects $\mathrm{BA}$. The explanations of this behaviour and its outcome assign large positive attributions to the position clue; this clue narrows the set of possible codes. 
The prediction explanation, in contrast, assigns a large negative attribution to the position clue; a clue of $1$ leaves two possible hidden codes, while a clue of $0$ would have ruled out all but one. The agent’s expected return is lower under the observed clue than it would be under a more informative one.

In the final state, the agent selects $\mathrm{AB}$ after receiving clues uniquely identifying the hidden code. The features corresponding to the previous guess and its feedback receive high attribution across all three explanations: they narrow down the optimal action (behaviour), which yields the high return received (outcome), and signal that the agent is in a state with maximal expected return (prediction).

\subsection{Minesweeper}
\label{subsec:minesweeper}

\begin{figure}[t]
\begin{center}
\centerline{\includegraphics[width=0.7\columnwidth]{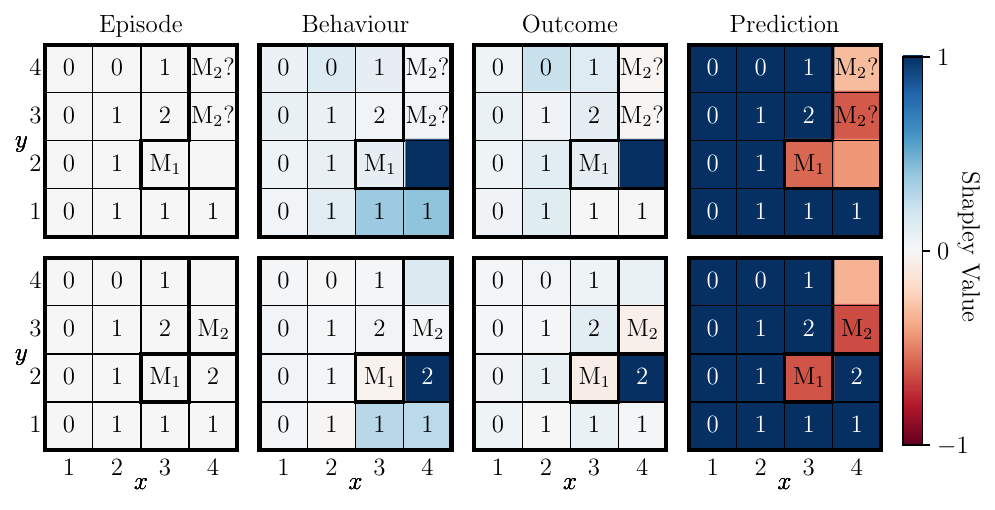}}
\vspace{-1em}
\caption{Shapley value attributions for two successive states in \texttt{Minesweeper} (top to bottom). First column: episode progression. Second column: feature contributions to the probability of selecting the optimal action taken (behaviour). Third column: contributions to the agent’s expected return (outcome). Fourth column: contributions to the agent’s estimated expected return (prediction). Mines are hidden from the agent; markers ``\textsc{M}$_1$" and ``\textsc{M}$_2$" are shown for reference. Shapley values are scaled to lie between $-1$ and $1$.}
\vspace{-2em}
\label{fig:minesweeper_sverl}
\end{center}
\end{figure}


\texttt{Minesweeper} is a puzzle in which the player must open squares on a grid while avoiding hidden mines. Selecting a square reveals either a mine (ending the game) or a number indicating how many adjacent squares contain mines. If a square with zero adjacent mines is revealed, its neighbouring squares are recursively opened until nonzero borders are reached. The game ends when either all safe squares have been uncovered or a mine is selected. Our formulation is played on a $4 \times 4$ grid, and each episode begins with two mines randomly placed and all squares unopened. We model the game as a Markov decision process with approximately 175{,}000 unique states: the agent selects an unopened square at each decision stage, receiving a reward of $-1$ if it reveals a mine (terminating the episode), and $0$ otherwise. Each state is represented by the visible game board, encoded as 16 features---one for each square---taking values \texttt{unopened}, or 0, 1 or 2 indicating the number of adjacent mines.

We use SVERL to analyse an agent in \texttt{Minesweeper} to show how explanations of behaviour, outcomes, and predictions scale to the feasible limits of exact Shapley value computation. We explain a policy trained using Q-learning \citep{Watkins1992}, with the expected return from outcome characteristic function estimated via Monte Carlo rollouts. \cref{fig:minesweeper_sverl} shows SVERL's explanations for states from two sequential decision stages.

At the first decision stage, the agent opens square $(4, 2)$; at the second, it opens square $(4, 4)$. In both stages, the behaviour and outcome explanations assign the highest attribution to the feature at $(4, 2)$. Many features help identify the location of the first mine (\textsc{M}$_1$), but only the clue revealed at square $(4, 2)$ uniquely determines the location of the second mine (\textsc{M}$_2$). SVERL captures this by assigning the highest attribution to that feature, reflecting its role in determining the agent's optimal action and the resulting expected return.

The prediction explanation shows a consistent pattern: positive attributions to opened squares and negative attributions to unopened ones. States with more unopened squares often carry greater uncertainty about mine locations and thus often have lower expected returns. SVERL's explanation of predictions reflects this relationship, where open squares, associated with greater certainty, lead to higher return estimates.

%% file: sections/8_discussion.tex
\section{Understanding and Interpreting SVERL}
\label{sec:discussion}

SVERL provides a lens through which we can understand how individual feature values contribute to an agent’s behaviour, outcomes, and predictions. These insights go beyond what the policy or expected return reveal alone. While many of the insights from our examples may seem intuitive in hindsight, they were not obvious when observing only the agent’s actions or value function. Even in well-understood domains, SVERL revealed how specific features shaped decision-making, outcomes, and predictions, offering a more detailed understanding of reinforcement learning.

In this section, we reflect on the insights provided by SVERL. We first examine how the three explanation types offer complementary perspectives that together form a richer picture of agent-environment interactions. We then discuss interpreting these explanations in practice, including common pitfalls and the risk of misunderstanding.

\subsection{SVERL's Complementary Insights}
\label{subsec:comparingexplanations}

Each explanation in the proposed framework captures a distinct aspect of reinforcement learning. Explanations of behaviour reveal how features influence the agent’s action choices. Explanations of outcomes reveal how features influence the consequences of the agent's behaviour, such as expected return. Explanations of predictions reveal how features influence the estimation of outcomes, such as estimated expected return. While each perspective offers insight, considering them together provides a more complete understanding of agent-environment interactions.

Consider the initial \texttt{Mastermind} state in \cref{fig:mastermind_sverl}. Here, the behaviour and prediction explanations assign importance to the same features with equal magnitudes but opposite signs, reflecting that the information these features convey about the underlying state simultaneously increases the likelihood of the agent's action while decreasing its estimate of expected return. In contrast, the outcome explanation attributes no importance to these features, illustrating a subtle but important point: features can influence behaviour and be used to predict outcomes without affecting the outcomes themselves. By comparison, the same features influence all three explanations in the final state. They identify the optimal action, which leads to the highest return---aligning behaviour, outcomes, and predictions in a single coherent interpretation.

These relationships illustrate how SVERL’s explanations, though distinct in focus, can meaningfully interact. They highlight how features play layered roles in shaping decisions, outcomes, and estimates.

\subsection{Interpreting SVERL}
\label{subsec:interpretingsverl}

As with any attribution method, interpreting SVERL requires care. Shapley values quantify the contribution of each feature to a particular explanatory target, such as an action probability, expected return, or value estimate. They do not reveal causal mechanisms or the internal reasoning behind the agent’s choices. Human users often construct explanations based on these attributions, but doing so can lead to overinterpretation or misplaced causal inference. Beyond their mathematical foundations, for Shapley-based explanations to be useful in practice, insights from cognitive science and psychology can help bridge the gap between statistical association and human understanding. SVERL provides a principled basis for attribution, but further analysis can help ensure that its outputs support meaningful and useful explanations in particular domains.

\begin{figure}[!t]
    \centering
    \includegraphics[width=0.6\columnwidth]{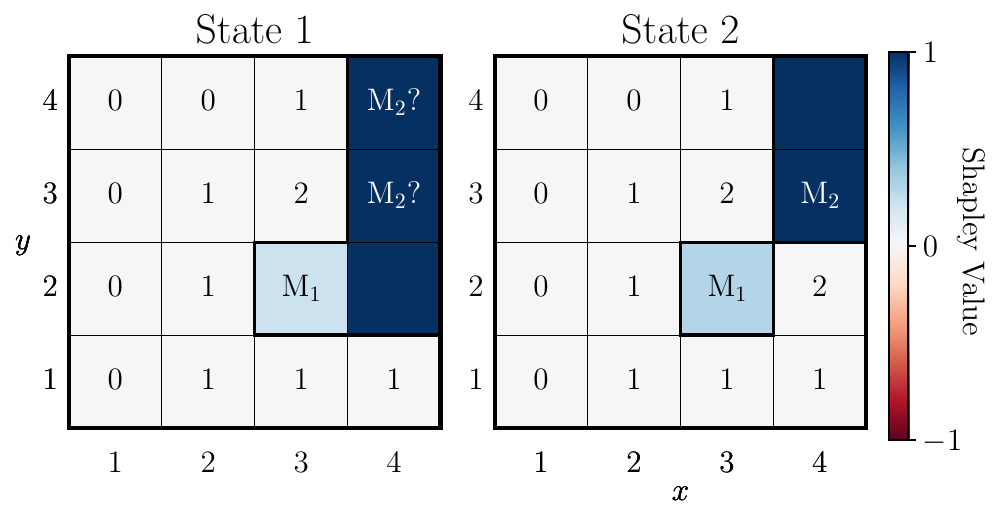}
    \caption{Feature contributions to the probability of opening each available square in the two states of \texttt{Minesweeper} shown in \cref{fig:minesweeper_sverl}. The colour of each square shows its Shapley value (scaled between $-1$ and $1$) for contributing to the probability of opening that same square.}
\label{fig:minesweeper_policy}
\end{figure}

Consider the example in \texttt{Minesweeper}. In \cref{fig:minesweeper_sverl}, SVERL assigns negative outcome attributions to unopened squares that contain mines in the second state. One might conclude that observing these features lowers expected return because it increases the likelihood that the agent opens them. This is a reasonable hypothesis, but it must be validated. \cref{fig:minesweeper_policy} shows the contribution of each square to the probability of the agent opening that square. In both states, the unopened squares contribute positively to the probability of being opened. This supports the hypothesis, though it was not a guaranteed conclusion from the original explanation alone.

Misunderstandings can also arise when explaining contributions by focusing too narrowly on individual marginal terms within the Shapley summation. An example of this is the misinterpretation of the $x$ and $y$ contributions in the \texttt{Taxi} domain by \citet{Beechey2023}, where these terms were accidentally swapped (corrected in \cref{sec:additionalexamples}). While the qualitative discussion was not entirely incorrect, it misrepresented the full contributions by considering only a subset of feature interactions, overlooking that Shapley values are means over all possible feature coalitions. This highlights the importance of interpreting Shapley-based explanations in light of the full combinatorial structure they represent.

Future work should explore how to effectively support users in interpreting SVERL. This includes developing tools for presenting explanations, educating users on how to interpret feature attributions responsibly, and providing methods for testing and validating the hypotheses they generate. Crucially, these efforts should be grounded in actual use cases and evaluated through user studies to ensure that SVERL contributes meaningfully to understanding and decision-making in practice.

%% file: sections/9_related_work.tex
\section{Related Work} 
\label{sec:relatedwork}

We now situate the proposed framework within the broader landscape of explainable reinforcement learning. We first review prior work that has applied Shapley values to analyse reinforcement learning agents, then discuss other lines of research in explainability.

\subsection{Previous Use of Shapley Values in Reinforcement Learning}
\label{subsec:shapleyinrl}

While Shapley values have previously been used to explain reinforcement learning agents, this work is the first to study them mathematically in this setting, focusing on the validity and meaning of the resulting explanations.

Several prior works analyse the value function using SHAP \citep{Lundberg2017}, a Python package that uses marginal characteristic functions to compute Shapley values for models \citep{Zhang2020, Schreiber2021, Zhang2022}. Applied to the value function, this setup corresponds to a marginal version of SVERL's prediction explanation, which assumes feature independence. However, as shown throughout this paper, prediction concerns the estimation of expected return, not how features influence the agent’s behaviour or the expected return the agent actually receives. Interpreting such explanations as relating to behaviour risks misunderstanding the agent’s interaction with its environment.

This distinction is illustrated in the \texttt{Tic-Tac-Toe} example of \cref{fig:tic_tac_toe_perf_value} in \cref{subsec:contrast}. In that example, the outcome explanation intuitively attributes responsibility to the opponent’s marked squares. Prediction, however, assigns zero contribution to all features because all states lead to a draw with expected return zero. There is no statistical association between feature values and return estimates; predictions can identify informative features for estimating return, but it does not explain how those features shape behaviour or outcomes.

Other works use SHAP to explain agent behaviour by analysing the policy \citep{Carbone2020, He2020, Wang2020, Liessner2021, Lover2021, Remman2021, Theumer2022}. These approaches attribute a deterministic action output by the policy model to individual input features. In continuous domains, this corresponds to SVERL’s continuous behaviour explanation (\cref{subsec:continuousactions}), where the Gaussian policy’s mean is treated as the selected action and the variance is fixed at zero. In discrete domains, the characteristic function---which averages over actions---can yield non-discrete outputs for unknown features, which do not correspond to any valid action. In contrast, SVERL provides principled behaviour explanations for discrete action spaces by attributing responsibility via action probabilities (\cref{subsec:policycharacteristic}).

Another line of work uses SHAP to explain stochastic policies in discrete action spaces by attributing feature importance to action probabilities \citep{Rizzo2019}. This aligns with SVERL’s behaviour explanation using the marginal characteristic function. However, the distribution over which the characteristic function is computed is left unspecified, likely resulting in misinterpretations of the resulting attributions. In contrast, SVERL uses the steady-state distribution to produce principled and interpretable attributions.

Finally, \citet{Waldchen2022} compute Shapley values by learning characteristic functions through training the policy and the value function with masked features. This approach restricts the number of masked features to stabilise learning, undermining the Shapley principle of evaluating all possible feature subsets. This is an important limitation because a feature’s importance may only emerge when enough other features are removed. This limitation prevents the method from satisfying the Shapley axioms and producing principled attributions, unlike SVERL, which is axiomatically sound.

\subsection{Other Approaches to Explainable Reinforcement Learning}
\label{subsec:otherxrl}

Feature importance methods are a common approach to explaining reinforcement learning agents, aiming to quantify how state features influence agent behaviour. These methods are particularly popular in visual domains. In addition to Shapley-based approaches, common alternatives include perturbing inputs and observing changes in outputs~\citep{Greydanus2018, Puri2019} or computing input gradients~\citep{Wang2016}. However, these methods lack the theoretical guarantees provided by Shapley values. For example, some can produce inconsistent attributions across models with identical outputs \citep{Sundararajan2017}. In contrast, SVERL is grounded in the Shapley framework, ensuring consistency and offering a principled, model-agnostic foundation for attributing importance.

Another prominent direction replaces complex models with simpler, interpretable ones, such as decision trees~\citep{Silva2020, Topin2021}, or trains surrogate models that approximate the behaviour of a trained agent~\citep{Bastani2018, Jhunjhunwala2019, Bewley2021}. These methods can improve transparency but often trade off expressivity, with surrogate models failing to represent the original agent faithfully. Other work has sought to better balance interpretability with fidelity by increasing model complexity~\citep{Hein2018, Landajuela2021}, though this inevitably increases cognitive load for users~\citep{Dodge2021}. SVERL avoids this trade-off entirely by explaining the behaviour and outcomes of any given agent directly, without retraining or simplification.

Other approaches analyse MDP structure and policies more broadly. Some leverage knowledge of transition dynamics to simulate counterfactuals~\citep{Rupprecht2019}, identify causal influences~\citep{Madumal2020}, or estimate task completion probabilities~\citep{Cruz2019, Cruz2023}. Others rely on assumptions about reward decomposability to trace how different objectives contribute to value estimates~\citep{Juozapaitis2019} or identify critical states where incorrect decisions would lead to poor performance~\citep{Amir2018, Huang2018}. These methods can offer useful insights but depend on explicit assumptions about environment models or reward structure, or are limited in scope, highlighting states without explaining why actions are chosen. By contrast, SVERL provides general-purpose, instance-level behaviour explanations without making such assumptions.

%% file: sections/10_future_work.tex
\section{Limitations and Future Work}
\label{sec:futurework}

This paper has focused on establishing the theoretical foundations of SVERL as a unified framework for explaining various elements of reinforcement learning agents. There is significant scope for extension. In this section, we highlight several promising directions that build on the foundations we have presented here, from practical challenges to broader theoretical developments.

\subsection{Explaining Agent-Environment Interactions More Broadly}

We have focused on explaining agents at specific states. Understanding behaviour broadly often involves reasoning about trends across many interactions. While one could present the Shapley values for many individual states and actions, doing so imposes a high cognitive load and hinders comprehension~\citep{Covert2021}. Here, we explore how SVERL could be extended to summarise explanations across states or actions to provide broader insights.

A natural idea is to aggregate SVERL's explanations for individual states or actions by taking their expectation under some distribution $\D$. For example, given a Shapley value from a behaviour explanation $\phi_i(\tilde\pi_s^a)$ for action $a$ at state $s$, we compute:
\begin{equation}
\underset{(s, a) \sim \D}{\mathbb{E}}[\phi_i(\tilde\pi_s^a)],
\end{equation}
as a general framework for summarising feature contributions. While we illustrate this approach for explanations of behaviour, the same approach applies to outcomes and predictions. The choice of distribution $\D$ determines the kind of insight obtained. It could reflect, for example, the full range of the agent’s experience, focus on particularly salient states, or highlight trends conditional on specific feature values. 

Under this framework, one might explore many directions. We now consider two natural first ideas: aggregating explanations across states using the steady-state distribution and aggregating across actions using the policy. Each may offer different perspectives on how features influence behaviour or outcomes broadly.

Let $\mathcal{D} = p^\pi(s)$, the steady-state distribution under policy $\pi$. Then, for behaviour explanations, we compute:
\begin{equation} \nonumber
    \underset{s \sim p^\pi(s)}{\mathbb{E}}[\phi_i(\tilde\pi_s^a)],
\end{equation}
which represents the mean contribution of feature $i$ to the probability of action $a$ across the agent's experience, weighting each feature contribution by the probability that it is observed. However, perhaps counterintuitively, this quantity is always zero.

\begin{proposition}\label{prop:globalshapley}
    For any policy $\pi$, the expected contribution of any feature to the probability of action $a$ under the steady-state distribution is zero:
    \begin{equation} \nonumber
        \underset{s \sim p^\pi(s)}{\mathbb{E}}[\phi_i(\tilde\pi_s^a)] = 0.
    \end{equation}
\end{proposition}

This result follows from the efficiency axiom: Shapley values explain deviations from a baseline, which is defined as the expected action probability over $p^\pi(s)$. On average, these deviations must sum to zero over $p^\pi(s)$. A similar argument applies to predictions, as stated below.

\begin{corollary}
    For any policy $\pi$, the expected contribution of any feature to the estimate of expected return under the steady-state distribution is zero:
    \begin{equation} \nonumber
        \underset{s \sim p^\pi(s)}{\mathbb{E}}[\phi_i(\hat{v}^\pi_s)] = 0.
    \end{equation}
\end{corollary}

In both cases, while local explanations offer insight into individual decisions or predictions, their steady-state averages provide no signal. This does not rule out the use of other state distributions. Alternative choices for $\D$ may yield meaningful summaries, making this a promising direction for future work.

In contrast, outcome explanations can yield non-zero expectations:
\begin{equation} \nonumber
    \underset{s \sim p^\pi(s)}{\mathbb{E}}[\phi_i(\tilde{v}^\pi_s)] \neq 0.
\end{equation}
Here, the baseline is not defined as an expectation over $p^\pi(s)$, so feature attributions retain signal when averaged. However, whether this expectation provides useful insight remains unclear. We leave a full analysis of its interpretability to future work.

As another example, let $\D = \pi(s, a)$, the policy’s action distribution at a given state. Then, for behaviour explanations, we compute:
\begin{equation} \nonumber
    \underset{a \sim \pi(s, a)}{\mathbb{E}}[\phi_i(\tilde\pi_s^a)],
\end{equation}
which represents the mean contribution of feature $i$ to the probabilities of actions in state $s$, weighting each feature contribution by the probability that the action is chosen. One could interpret this concept as capturing the general influence of a feature value on behaviour across actions at a given state. By the linearity of Shapley values, this quantity is itself a Shapley value for a new coalitional game with characteristic function defined as the expected action probability:
\begin{equation} \nonumber
    \underset{a \sim \pi(s, a)}{\mathbb{E}}[\phi_i(\tilde\pi_s^a)] = \phi_i\left(\underset{a \sim \pi(s, a)}{\mathbb{E}}[\tilde\pi_s^a]\right).
\end{equation}
Its interpretation remains unclear, but this connection to cooperative game theory offers a promising direction for investigating what insight such a summary explanation provides, and whether it meaningfully captures a feature’s importance across the agent’s behaviour.

\subsection{Approximating SVERL}

Computing SVERL’s explanations is computationally demanding. Each characteristic function involves an expectation over the entire state space, and each Shapley value is defined as a sum over all $2^{|\F|}$ subsets of features, where $\F$ is the full set of features. These requirements quickly become intractable as the number of features or states grows. As a result, practical applications of SVERL in large domains require approximation techniques.

Approximation techniques have been widely studied in the supervised learning setting, and it is only natural to expect that many of these methods could be extended to SVERL. Here, we adapt one such method, the Monte Carlo sampling techniques in \cref{subsec:approximating_sv}, to approximate SVERL’s explanations of behaviour, outcomes, and predictions. We present this theoretical outline to demonstrate that SVERL can be approximated in practice, leaving its full treatment and discussion of alternative methods to future work.

We begin with explanations of behaviour and predictions. We propose approximating their characteristic functions, defined in \cref{eq:policychar,eq:valuechar}, using Monte Carlo sampling:
\begin{align} \label{eq:approxpolicy}
    \tilde\pi_s^a(\C) &= \mathbb{E}[\pi(S, a) \mid S_\C = s_\C] = \lim_{n\to \infty} \frac{1}{n} \sum_{s' \sim p^\pi(\cdot \mid S_\C = s_\C)} \pi(s', a), \\
    \hat{v}^\pi_s(\C) &= \mathbb{E}[\hat{v}^\pi(S) \mid S_\C = s_\C] = \lim_{n\to \infty} \frac{1}{n} \sum_{s' \sim p^\pi(\cdot \mid S_\C = s_\C)} \hat{v}^\pi(s').
\end{align}
Paralleling the approach in supervised learning, we combine these approximations with the permutation-based Shapley value definition in \cref{eq:svperm}, yielding a single Monte Carlo sampling method:
\begin{align*}
    \phi_i(\tilde\pi_s^a) &= \lim_{n\to \infty} \frac{1}{n} \sum_{\substack{
        s' \sim p^\pi(\cdot \mid S_{\C \cup \{i\}} = s_{\C \cup \{i\}}) \\
        s'' \sim p^\pi(\cdot \mid S_\C = s_\C)
    }} \left[\pi(s', a) - \pi(s'', a)\right], \\
    \phi_i(\hat{v}^\pi_s) &= \lim_{n\to \infty} \frac{1}{n} \sum_{\substack{
        s' \sim p^\pi(\cdot \mid S_{\C \cup \{i\}} = s_{\C \cup \{i\}}) \\
        s'' \sim p^\pi(\cdot \mid S_\C = s_\C)
    }} \left[\hat{v}^\pi(s') - \hat{v}^\pi(s'')\right],
\end{align*}
where $\C$ is the set of features that precede $i$ in ordering $O$ and $O$ is sampled uniformly from $\pi(\F)$, the set of all permutations of $\F$.

Outcome explanations require a different treatment. The characteristic function $\tilde{v}^\pi_s(\C)$, defined in \cref{eq:outcomechar}, is the expected return from state $s$ under a policy that follows the partially observed policy $\tilde\pi_s^a(\C)$ in $s$, and the fully observed policy $\pi$ elsewhere.

Because this characteristic is an expected return, it can be estimated using any standard reinforcement learning method. However, it depends on the discrete policy characteristic $\tilde\pi_s^a(\C)$, which must also be approximated. We propose incorporating the sampling-based approximation of $\tilde\pi_s^a(\C)$ from \cref{eq:approxpolicy} directly into the policy definition used within the outcome characteristic's expectation:
\begin{align*}
    \tilde{v}^\pi_s(\C) &= \mathbb{E}_{\hat\pi_\C}\left[\sum_{k=0}^\infty \gamma^k R_{t+k+1} \mid S_t = s\right],\\
    \text{where} \quad \hat\pi_\C(a_t \mid s_t) &=
    \begin{cases}
        \pi(a_t \mid s'), & \text{with } s' \sim p^\pi(\cdot \mid s_\C) \quad \text{if } s_t = s, \\
        \pi(a_t \mid s_t), & \text{otherwise.}
    \end{cases}
\end{align*}
This defines a new policy $\hat\pi_\C(a_t \mid s_t)$ that, when visiting $s$, selects actions based on a sampled state $s'$ consistent with the observed features: $s_\C = s'_\C$, and follows the fully observed policy $\pi$ elsewhere. By rewriting the outcome characteristic in this equivalent form, we embed the approximation of the discrete policy characteristic $\tilde\pi_s^a(\C)$ into the return estimation itself, allowing $\tilde{v}^\pi_s(\C)$ to be computed directly using standard reinforcement learning methods.

Finally, with $\tilde{v}^\pi_s(\C)$ approximated in this way, the Shapley values for outcomes can be computed using the usual permutation-based sum:
\begin{equation}\nonumber
    \phi_i(\tilde{v}^\pi_s) = \lim_{n\to \infty} \frac{1}{n} \sum_{O \sim \pi(\F)} \left[\tilde{v}^\pi_s(\C \cup \{i\}) - \tilde{v}^\pi_s(\C)\right].
\end{equation}

All three explanations---behaviour, outcomes, and predictions---require access to the conditional steady-state distribution $p^{\pi}(S \mid S_\C = s_\C)$. This distribution must be learned in large domains, presenting an important direction for future work. One promising starting point is the parametric approach introduced by \citet{Frye2020}, which uses variational inference to learn conditional data distributions from a dataset. A natural adaptation to the reinforcement learning setting would be to use this method to learn the conditional steady-state distribution from a replay buffer of agent experience.

Here we have presented an outline for approximating SVERL by adapting Monte Carlo techniques from supervised learning to the reinforcement learning setting, without investigating empirical performance or efficiency. We encourage a full treatment of these questions in subsequent research. Exploring other approximation methods developed in supervised learning may also prove fruitful. For example, \citet{Frye2020} propose learning parametric models of the characteristic function directly, and \citet{Jethani2021} amortises Shapley value approximations across inputs by training parametric models to predict them. These approaches are promising future avenues to enable scalable or efficient approximations of SVERL.

\subsection{Using Predictions to Inform Behaviour}
\label{subsec:usingvalueestimation}

SVERL's explanations of predictions reveal how feature values influence the agent's estimated expected return. As shown in the examples in \cref{sec:examples}, feature values in a given state that also appear in high-return states tend to receive positive attributions. In contrast, feature values associated with lower-return states tend to receive negative attributions. This suggests that these explanations are useful for highlighting which aspects of the environment correlate with successful estimated outcomes. One intriguing possibility is to use this information to inform behaviour---for example, encouraging actions that lead to positively attributed features, or identifying negatively attributed features that should be changed to improve expected return. The additional example in \cref{sec:additionalexamples} illustrates this idea concretely: in a dice-rolling domain, each die is treated as a feature and receives a negative attribution if and only if the optimal policy re-rolls the corresponding die to change its value. How to effectively incorporate these signals into decision-making or policy optimisation is a compelling question, and one we believe represents an important direction for future work.

\subsection{Connecting SVERL to Broader Attribution Methods in Cooperative Game Theory}

SVERL provides a principled framework for explaining reinforcement learning agents using Shapley values. It introduces coalitional games tailored to reinforcement learning by defining characteristic functions for behaviour, outcomes, and predictions. Because the framework inherits the structure and axiomatic foundations of classic cooperative game theory, it can naturally accommodate alternative definitions, theoretical developments, and methodological advances from the broader literature.

To illustrate the kinds of opportunities this generality creates, we highlight two possible extensions. First, \citet{Frye2020a} propose computing Shapley values under a causal model, modifying the Shapley sum to account for known dependencies between features. For example, if one feature is a deterministic causal ancestor of another, standard Shapley values may split credit between them; a causal approach would attribute responsibility to the source feature. Extending SVERL in this way could yield explanations that more faithfully reflect the causal structure of the environment or agent behaviour.

Another possibility is to explore alternative solution concepts from cooperative game theory. For instance, Banzhaf values \citep{Banzhaf1964} define a different notion of attribution based on a distinct set of axioms and may offer complementary insights when applied to SVERL's characteristic functions. These examples demonstrate only two possible ways broader theoretical developments in attribution research can be incorporated into reinforcement learning using SVERL. Exploring these connections provides a promising avenue for future work.

%% file: sections/11_conclusion.tex
\section{Conclusion}
\label{sec:conclusion}

Reinforcement learning agents can surpass human performance, but the how and why of their interactions with their environments remain difficult to interpret. In this work, we introduced Shapley Values for Explaining Reinforcement Learning (SVERL), a theoretical framework for explaining agent behaviour, outcomes, and predictions through the influence of features at specific states. These explanations are grounded in the well-established cooperative game-theoretic foundations of Shapley values, providing a family of theoretically principled methods for attribution.

The three explanatory elements---behaviour, outcomes, and predictions---offer distinct but complementary perspectives on agent-environment interaction. Their interdependence reveals how features can play layered roles in shaping actions, consequences, and estimates, highlighting the complexity of attribution in sequential settings and the value of unified explanatory frameworks.

SVERL formalises feature influence using Shapley values, offering principled, axiomatic explanations. However, these quantify statistical contributions rather than causal mechanisms, and thus demand careful interpretation. Translating mathematically grounded explanations into practical, trustworthy insights will require new tools, clearer user guidance, and closer alignment with findings from cognitive science.

Scaling SVERL introduces both computational and cognitive challenges. While Monte Carlo approximations can make attribution tractable, and aggregating local attributions into behavioural summaries may reduce interpretive burden, these strategies still require empirical validation and methodological refinement to support real-world deployment.

More broadly, this work situates the explanation of reinforcement learning within a rich theoretical landscape. The cooperative game-theoretic foundation invites a wide range of future extensions, including alternative characteristic functions, new solution concepts, and methodological advances, all within a common, principled language.

By establishing a theoretical basis for explaining reinforcement learning agents, SVERL contributes to the development of reinforcement learning systems that are not only powerful, but also transparent, trustworthy, and aligned with human understanding.

%% file: sections/acknowledgements.tex
\acks{This work was supported by the Engineering and Physical Sciences Research Council (EPSRC) [grant number EP/X025470/1], UKRI Centre for Doctoral Training in Accountable, Responsible and Transparent AI (ART-AI) [EP/S023437/1], the EPSRC Centre for Doctoral Training in Digital Entertainment (CDE) [EP/L016540/1],  and the University of Bath. This research made use of Hex, the GPU Cloud in the Department of Computer Science at the University of Bath.
}

%% file: sections/appendix/proofs.tex
\section{Proofs of Propositions and Results}
\label{sec:proofs}

We present each result and its associated proof in full detail.

\paragraph{Proposition \ref{prop:policyshapley}} \textit{
    The Shapley values $\phi_1(\tilde\pi_s^a), \dots, \phi_n(\tilde\pi_s^a)$ for the discrete policy game $(\F, \tilde\pi_s^a)$ are the unique allocation of the action probability $\pi(s, a)$ across the features of state $s$ that satisfies the four axioms of fair credit assignment: efficiency, linearity, symmetry, and nullity.
} \\

\begin{proof}
    For a finite set of features $\F$, the characteristic function $\tilde\pi_s^a: 2^{\F} \to \mathbb{R}$ has well-defined values for all coalitions $\C \subseteq \F$. Therefore, the policy game is well-defined. By Shapley’s theorem \citep{Shapley1953}, the Shapley values $\phi_1(\tilde\pi_s^a), \dots, \phi_n(\tilde\pi_s^a)$ are the unique value allocation satisfying efficiency, linearity, symmetry, and nullity.
\end{proof}

\paragraph{Proposition \ref{prop:policychar}} \textit{
    When comparing policies using mean squared error between the probabilities of selecting action $a$, the policy $\mu$ is the policy that deviates the least in expectation from the policy $\pi$:
    \begin{equation}
        \mu = \argmin_g \mathbb{E}\left[\lVert \pi(S, a) - g(S_\C, a) \rVert^2\right],
    \end{equation}
    where $g: \S_\C \rightarrow \Delta(\A)$ is also defined over the subset of features $\C$.
} \\

\begin{proof}
    We rearrange the expected squared error:
    \begin{align*}
        \mathbb{E}\left[\lVert \pi(S, a) - g(S_\C, a) \rVert^2\right] 
        &= \mathbb{E}_S\left[\lVert \pi(S, a) - \mathbb{E}[\pi(S, a)\,|\,S_\C] \rVert^2\right] \\
        &\quad + \mathbb{E}_{S_\C}\left[\lVert \mathbb{E}[\pi(S, a)\,|\,S_\C] - g(S_\C, a) \rVert^2\right].
    \end{align*}
    The first term is independent of $g$, and the second is minimised when $g(S_\C, a) = \mathbb{E}[\pi(S, a)\,|\,S_\C] = \mu(S_\C, a)$. Substituting $g = \mu$ gives:
    \begin{align*}
        \mathbb{E}\left[\lVert \pi(S, a) - \mu(S_\C, a) \rVert^2\right] 
        &= \mathbb{E}_S\left[\lVert \pi(S, a) - \mathbb{E}[\pi(S, a)\,|\,S_\C] \rVert^2\right] \\
        &\quad + \mathbb{E}_{S_\C}\left[\lVert \mathbb{E}[\pi(S, a)\,|\,S_\C] - \mathbb{E}[\pi(S, a)\,|\,S_\C] \rVert^2\right] \\
        &= \mathbb{E}_S\left[\lVert \pi(S, a) - \mathbb{E}[\pi(S, a)\,|\,S_\C] \rVert^2\right],
    \end{align*}
    confirming that $\mu$ achieves the minimum.
\end{proof}

\paragraph{Proposition \ref{prop:continuouspolicyshapley}} \textit{
    The Shapley values $\phi_1(\tilde\mu_s), \dots, \phi_n(\tilde\mu_s)$ for the continuous policy game $(\F, \tilde\mu_s)$ are the unique allocation of the expected action $\mu(s)$ across the features of state $s$ that satisfies the four axioms of fair credit assignment: efficiency, linearity, symmetry, and nullity.
} \\

\begin{proof}
    For a finite set of features $\F$, the characteristic function $\tilde\mu_s: 2^{\F} \to \mathbb{R}$ has well-defined values for all coalitions $\C \subseteq \F$. Therefore, the continuous policy game is well-defined. By Shapley’s theorem \citep{Shapley1953}, the Shapley values $\phi_1(\tilde\mu_s), \dots, \phi_n(\tilde\mu_s)$ are the unique value allocation satisfying efficiency, linearity, symmetry, and nullity.
\end{proof}

\paragraph{Proposition \ref{prop:continuouspolicychar}} \textit{
    When comparing mean functions using mean squared error, the function $\omega$ deviates the least in expectation from the mean $\mu$ of the full policy $\pi$:
    \begin{equation}
        \omega = \argmin_{g: \S_\C \rightarrow \mathbb{R}} \mathbb{E}\left[\lVert \mu(S) - g(S_\C) \rVert^2\right].
    \end{equation}
} \\

\begin{proof}
    We rearrange the expected squared error:
    \begin{align*}
        \mathbb{E}\left[\lVert \mu(S) - g(S_\C) \rVert^2\right] 
        &= \mathbb{E}_S\left[\lVert \mu(S) - \mathbb{E}[\mu(S)\,|\,S_\C] \rVert^2\right] \\
        &\quad + \mathbb{E}_{S_\C}\left[\lVert \mathbb{E}[\mu(S)\,|\,S_\C] - g(S_\C) \rVert^2\right].
    \end{align*}
    The first term is independent of $g$, and the second is minimised when $g(S_\C) = \mathbb{E}[\mu(S)\,|\,S_\C] = \omega(S_\C)$. Substituting $g = \omega$ gives:
    \begin{align*}
        \mathbb{E}\left[\lVert \mu(S) - \omega(S_\C) \rVert^2\right] 
        &= \mathbb{E}_S\left[\lVert \mu(S) - \mathbb{E}[\mu(S)\,|\,S_\C] \rVert^2\right] \\
        &\quad + \mathbb{E}_{S_\C}\left[\lVert \mathbb{E}[\mu(S)\,|\,S_\C] - \mathbb{E}[\mu(S)\,|\,S_\C] \rVert^2\right] \\
        &= \mathbb{E}_S\left[\lVert \mu(S) - \mathbb{E}[\mu(S)\,|\,S_\C] \rVert^2\right],
    \end{align*}
    confirming that $\omega$ achieves the minimum.
\end{proof}

\paragraph{Proposition \ref{prop:outcomeshapley}} \textit{
    The Shapley values $\phi_1(\tilde{v}^\pi_s), \dots, \phi_n(\tilde{v}^\pi_s)$ for the outcome game $(\F, \tilde{v}^\pi_s)$ are the unique allocation of the expected return $v^\pi(s)$ across the features of state~$s$ that satisfies the four axioms of fair credit assignment: efficiency, linearity, symmetry, and nullity.
} \\

\begin{proof}
    For a finite set of features $\F$, the characteristic function $\tilde{v}^\pi_s(\C): 2^{\F} \to \mathbb{R}$ has well-defined values for all coalitions $\C \subseteq \F$. Therefore, the outcome game is well-defined. By Shapley’s theorem \citep{Shapley1953}, the Shapley values $\phi_1(\tilde{v}^\pi_s), \dots, \phi_n(\tilde{v}^\pi_s)$ are the unique value allocation satisfying efficiency, linearity, symmetry, and nullity.
\end{proof}

\paragraph{Proposition \ref{prop:valueshapley}} \textit{
    The Shapley values $\phi_1(\hat{v}^\pi_s), \dots, \phi_n(\hat{v}^\pi_s)$ for the prediction game $(\F, \hat{v}^\pi_s)$ are the unique allocation of the predicted expected return $\hat{v}^\pi(s)$ across the features of state~$s$ that satisfies the four axioms of fair credit assignment: efficiency, linearity, symmetry, and nullity.
} \\

\begin{proof}
    For a finite set of features $\F$, the characteristic function $\hat{v}^\pi_s: 2^{\F} \to \mathbb{R}$ has well-defined values for all coalitions $\C \subseteq \F$. Therefore, the prediction game is well-defined. By Shapley’s theorem \citep{Shapley1953}, the Shapley values $\phi_1(\hat{v}^\pi_s), \dots, \phi_n(\hat{v}^\pi_s)$ are the unique value allocation satisfying efficiency, linearity, symmetry, and nullity.
\end{proof}

\paragraph{Proposition \ref{prop:valuechar}} \textit{
    When comparing expected-return estimates using mean squared error, the function $\hat{u}^\pi$ is the closest in expectation to $\hat{v}^\pi$ among all functions defined over the subset of features $\C$:
    \begin{equation}\nonumber
        \hat{u}^\pi = \argmin_{g: \S_\C \rightarrow \mathbb{R}} \mathbb{E}\left[\lVert \hat{v}^\pi(S) - g(S_\C) \rVert^2\right].
    \end{equation}
} \\

\begin{proof}
    We rearrange the expected squared error:
    \begin{align*}
        \mathbb{E}\left[\lVert \hat{v}^\pi(S) - g(S_\C) \rVert^2\right] 
        &= \mathbb{E}_S\left[\lVert \hat{v}^\pi(S) - \mathbb{E}[\hat{v}^\pi(S)\,|\,S_\C] \rVert^2\right] \\
        &\quad + \mathbb{E}_{S_\C}\left[\lVert \mathbb{E}[\hat{v}^\pi(S)\,|\,S_\C] - g(S_\C) \rVert^2\right].
    \end{align*}
    The first term is independent of $g$, and the second is minimised when $g(S_\C) = \mathbb{E}[\hat{v}^\pi(S)\,|\,S_\C] = \hat{u}^\pi(S_\C)$. Substituting $g = \hat{u}^\pi$ gives:
    \begin{align*}
        \mathbb{E}\left[\lVert \hat{v}^\pi(S) - \hat{u}^\pi(S_\C) \rVert^2\right] 
        &= \mathbb{E}_S\left[\lVert \hat{v}^\pi(S) - \mathbb{E}[\hat{v}^\pi(S)\,|\,S_\C] \rVert^2\right] \\
        &\quad + \mathbb{E}_{S_\C}\left[\lVert \mathbb{E}[\hat{v}^\pi(S)\,|\,S_\C] - \mathbb{E}[\hat{v}^\pi(S)\,|\,S_\C] \rVert^2\right] \\
        &= \mathbb{E}_S\left[\lVert \hat{v}^\pi(S) - \mathbb{E}[\hat{v}^\pi(S)\,|\,S_\C] \rVert^2\right],
    \end{align*}
    confirming that $\hat{u}^\pi$ achieves the minimum.
\end{proof}

\paragraph{Proposition \ref{prop:globalshapley}} \textit{
    For any policy $\pi$, the expected contribution of any feature to the probability of action $a$ under the steady-state distribution is zero:
    \begin{equation} \nonumber
        \underset{s \sim p^\pi(s)}{\mathbb{E}}[\phi_i(\tilde\pi_s^a)] = 0.
    \end{equation}
} \\

\begin{proof}
    Begin by expanding the definition of the Shapley value and applying the linearity of expectation:
    \begin{align*}
        \mathbb{E}_{p^\pi(s)}[\phi_i(\tilde\pi_s^a)] 
        &= \mathbb{E}_{p^\pi(s)}\left[\sum_{\C\subseteq \F \setminus \{i\}} \frac{|\C|!(|\F| - |\C| - 1)!}{|\F|!} \left[\tilde\pi_s^a(\C \cup \{i\}) - \tilde\pi_s^a(\C)\right]\right] \\
        &= \sum_{\C\subseteq \F \setminus \{i\}} \frac{|\C|!(|\F| - |\C| - 1)!}{|\F|!} \left[\mathbb{E}_{p^\pi(s)}[\tilde\pi_s^a(\C \cup \{i\})] - \mathbb{E}_{p^\pi(s)}[\tilde\pi_s^a(\C)]\right].
    \end{align*}

    To simplify each expectation term, recall that $\tilde\pi_s^a(\C)$ is defined as the conditional expectation of $\pi(s, a)$ given $S_\C = s_\C$, followed by marginalisation over $s_\C$:
    \begin{align*}
        \mathbb{E}_{p^\pi(s)}[\tilde\pi_s^a(\C)] 
        &= \mathbb{E}_{s_\C \sim p^\pi(s_\C)}\left[\mathbb{E}[\pi(S, a) \mid S_\C = s_\C]\right] 
        = \mathbb{E}_{p^\pi(s)}[\pi(S, a)].
    \end{align*}

    That is, marginalising the conditional expectation recovers the unconditional expectation. The same holds for $\C \cup \{i\}$, so:
    \begin{align*}
        \mathbb{E}_{p^\pi(s)}[\tilde\pi_s^a(\C \cup \{i\})] = \mathbb{E}_{p^\pi(s)}[\pi(S, a)].
    \end{align*}

    Substituting back, the two terms in the Shapley difference cancel:
    \begin{align*}
        \mathbb{E}_{p^\pi(s)}[\phi_i(\tilde\pi_s^a)] 
        &= \sum_{\C\subseteq \F \setminus \{i\}} \frac{|\C|!(|\F| - |\C| - 1)!}{|\F|!} \left[\mathbb{E}[\pi(S, a)] - \mathbb{E}[\pi(S, a)]\right] \\
        &= 0.
    \end{align*}
    This concludes the proof.
\end{proof}

%% file: sections/appendix/additional_examples.tex
\section{Additional Examples}
\label{sec:additionalexamples}

We present three examples to illustrate how our framework attributes behaviour, outcomes and predictions to features. We then present an additional example explaining all three. 


\begin{figure}
    \begin{subfigure}[b]{0.18\textwidth}
        \centering
        \begin{adjustbox}{valign=c}
            \includegraphics[width=\textwidth]{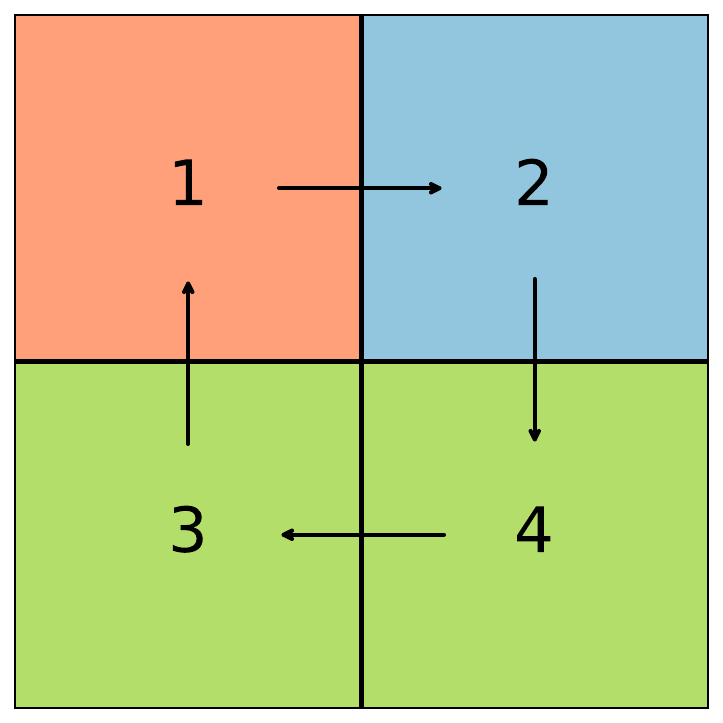}
        \end{adjustbox}
    \end{subfigure}
    \begin{subfigure}[b]{0.82\textwidth}
        \centering
        \begin{adjustbox}{valign=c}
            \renewcommand{\arraystretch}{1.15}
            \resizebox{\textwidth}{!}{%
            \begin{tabular}{cc|c|cccc|cc}
            \toprule
            \textbf{State ($s$)} & \textbf{Action ($a$)} & $p^\pi(s)$ & $\tilde\pi_s^a(\text{Both})$ & $\tilde\pi_s^a(\text{I})$ & $\tilde\pi_s^a(\text{C})$ & $\tilde\pi_s^a(\emptyset)$ & $\phi_{\text{I}}$ & $\phi_{\text{C}}$ \\
            \midrule
            \multirow{4}{*}{$(1,\text{red})$} & N & \multirow{4}{*}{$0.25$} & $0$ & $0$ & $0$ & $0.25$ & $-0.125$ & $-0.125$ \\
                          & E &  & $1$ & $1$ & $1$ & $0.25$ & $0.375$ & $0.375$ \\
                          & S &  & $0$ & $0$ & $0$ & $0.25$ & $-0.125$ & $-0.125$ \\
                          & W &  & $0$ & $0$ & $0$ & $0.25$ & $-0.125$ & $-0.125$ \\
            \midrule
            \multirow{4}{*}{$(3,\text{green})$} & N & \multirow{4}{*}{$0.25$} & $1$ & $1$ & $0.5$ & $0.25$ & $0.625$ & $0.125$ \\
                          & E &  & $0$ & $0$ & $0$ & $0.25$ & $-0.125$ & $-0.125$ \\
                          & S &  & $0$ & $0$ & $0$ & $0.25$ & $-0.125$ & $-0.125$ \\
                          & W &  & $0$ & $0$ & $0.5$ & $0.25$ & $-0.375$ & $0.125$ \\
            \bottomrule
            \end{tabular}
            }
        \end{adjustbox}
    \end{subfigure}%
    \caption{Explaining behaviour in a $2\times2$ gridworld.
    }
    \label{fig:colourgridpolicy}
\end{figure}

\textbf{Explaining Behaviour.} Consider the $2\times2$ gridworld shown in \cref{fig:colourgridpolicy}, in which each state is defined by two features: index (I) and colour (C). The agent's policy ($\pi$) is to move clockwise through the grid. We analyse how the agent’s action probabilities in states $(1, \text{red})$ and $(3, \text{green})$ change as different features become known. These probabilities define SVERL's characteristic values for explaining behaviour, from which the Shapley values are computed, as shown in the figure. Consider first the state $(1, \text{red})$. When neither feature is known, the agent is equally likely to be in any state (by the steady-state distribution), and thus acts uniformly over the four actions. If the index becomes known to be $1$, or the colour becomes known to be red, the agent must be in state $(1, \text{red})$ and selects East with probability~1. Thus, both features fully resolve the agent’s uncertainty and have the same effect on the action probabilities. This is reflected in the Shapley values, which assign equal contributions to the index and colour. 
Next consider state $(3, \text{green})$. If the index becomes known to be $3$, the agent must be in state $(3, \text{green})$ and selects North with probability~1. If instead the colour becomes known to be green, the agent could be in either state $(3, \text{green})$ or $(4, \text{green})$, and hence selects North and West with equal probability. Thus, both features increase the probability of the correct action (North), the index increases this probability more, and the colour increases the probability of an incorrect action (West). 


\begin{figure}[!tb]
    \graphicspath{{./plots/}}
    \centering
    \begin{subfigure}[c]{\textwidth}
        \centering
        \begin{subfigure}[b]{0.35\textwidth}
            \centering
            \begin{adjustbox}{valign=c}
                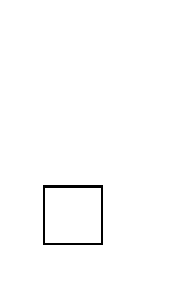
            \end{adjustbox}
        \end{subfigure}%
        \hfill
        \begin{subfigure}[c]{0.65\textwidth}
            \centering
            \renewcommand{\arraystretch}{1.15}
            \resizebox{\textwidth}{!}{%
            \begin{tabular}{cc|cccc}
            \toprule
            \textbf{State $(s)$} & \textbf{Action ($a$)} & $\tilde\pi_s^a(\text{Both})$ & $\tilde\pi_s^a(x)$ & $\tilde\pi_s^a(y)$ & $\tilde\pi_s^a(\emptyset)$ \\
            \midrule
            \multirow{4}{*}{1} & N & 0 & 0 & $2/3$ & $6/7$ \\
                               & E & 1 & 1 & $1/3$ & $1/7$ \\
                               & S & 0 & 0 & 0 & 0 \\
                               & W & 0 & 0 & 0 & 0 \\
            \midrule
            \multirow{4}{*}{2} & N & 1 & 1 & $2/3$ & $6/7$ \\
                               & E & 0 & 0 & $1/3$ & $1/7$ \\
                               & S & 0 & 0 & 0 & 0 \\
                               & W & 0 & 0 & 0 & 0 \\
            \bottomrule
            \end{tabular}
            }
        \end{subfigure}
    \end{subfigure}

    \vspace{0.5em}

    \begin{subfigure}[b]{\textwidth}
        \centering
        \resizebox{0.80\textwidth}{!}{%
        \begin{tabular}{c|c|cccc|cc}
        \toprule
        \textbf{State $(s)$} & $p^\pi(s)$ & $\tilde{v}^\pi_s(\text{Both})$ & $\tilde{v}^\pi_s(x)$ & $\tilde{v}^\pi_s(y)$ & $\tilde{v}^\pi_s(\emptyset)$ & $\phi_x$ & $\phi_y$ \\
        \midrule
        1 & $1/7$ & $6.00$ & $6.00$ & $4.00$ & $0.00$ & $4.00$ & $2.00$ \\
        2 & $2/7$ & $7.00$ & $7.00$ & $6.50$ & $6.83$ & $0.33$ & $-0.17$ \\
        \bottomrule
        \end{tabular}
        }
    \end{subfigure}

    \caption{Explaining expected return in a five-state gridworld. 
    }
    \label{fig:gwoutcome}
\end{figure}

\textbf{Explaining Outcomes.} Consider the five-state gridworld shown in \cref{fig:gwoutcome} where the agent begins uniformly at random in state~1 or~2, receiving a reward of $-1$ per action and an additional $+10$ upon reaching a goal state. Invalid actions incur a reward but leave the agent's position unchanged. The figure shows the optimal deterministic policy $\pi$: East from state~1 and North from all others. Each state is described by two features: its $x$- and $y$-coordinates. We analyse how the expected return from each state changes as these coordinates become known, owing to the agent’s behaviour in that state shifting accordingly. These returns define SVERL’s characteristic values for explaining outcomes, while the resulting Shapley values quantify the influence of each feature. Both are shown in the bottom table of \cref{fig:gwoutcome}; the top-right table shows the corresponding action probabilities under different subsets of known features.

Consider state~1. When neither feature is known, the agent has probability $1/7$ of being in state~1 (with optimal action East) and probability $2/7$ of being in states~2, 3, and~4 (all with optimal action North). Therefore, the agent selects North with probability $6/7$ and East with $1/7$, yielding an expected return of $0$ from state~1. If the $x$-coordinate becomes known, the agent must be in state~1 and selects East, yielding a return of $6$. If instead the $y$-coordinate becomes known, the agent could be in state~1 or~2, with state~2 twice as likely, and so selects North with probability $2/3$ and East with $1/3$, giving an expected return of~$4$. This difference is reflected in the Shapley values in \cref{fig:gwoutcome}, which assign twice as much influence on expected return to the $x$-coordinate as the $y$-coordinate.

Now consider state~2. As in state~1, when no features are known, the agent follows the steady-state distribution, selecting North with probability $6/7$ and East with probability $1/7$, yielding an expected return of $6.83$. If the $x$-coordinate becomes known, the agent must be in states~2,~3, or~4 (all with optimal action North). Therefore, the agent deterministically selects North, increasing the return from state~2 to $7$. If instead the $y$-coordinate becomes known, the agent could be in state~1 or~2, with state~2 twice as likely, and so selects North with probability $2/3$ and East with $1/3$. This gives a lower expected return of $6.5$ because the likelihood of the optimal action (North) has decreased from $6/7$ to $2/3$. The Shapley values in \cref{fig:gwoutcome} reflect this: the $x$-coordinate contributes positively to expected return, while the $y$-coordinate has a negative influence.


\begin{table}[b]
    \centering
    \renewcommand{\arraystretch}{1.15}
    \resizebox{0.9\textwidth}{!}{%
    \begin{tabular}{c|c|cccc|cc}
    \toprule
    \textbf{State $(s)$} & $p^\pi(s)$ & $\hat{v}^\pi_s(\text{Both})$ & $\hat{v}^\pi_s(d_1)$ & $\hat{v}^\pi_s(d_2)$ & $\hat{v}^\pi_s(\emptyset)$ & $\phi_{d_1}$ & $\phi_{d_2}$ \\
    \midrule
    $(3, 6)$ & $0.024$ & $0.67$ & $0.45$ & $0.90$ & $0.66$ & $-0.22$ & $0.23$ \\
    $(1, 1)$ & $0.018$ & $0.36$ & $0.45$ & $0.45$ & $0.66$ & $-0.15$ & $-0.15$ \\
    \bottomrule
    \end{tabular}
    }
    \caption{Explaining predictions in a dice-rolling game. 
    }
    \label{tab:dicevalueestimation}
\end{table}

\textbf{Explaining Predictions.} Consider a dice-rolling game, in which each episode begins with a random roll of two six-sided dice. State is denoted $(d_1, d_2)$, where $d_1$ and $d_2$ are the first and second die values, respectively. The agent may re-roll either die, both dice, or neither. After each action, the episode ends with probability $0.5$, yielding a reward of $1$ if the sum of the dice is $10$ or more, and $0$ otherwise. The optimal policy keeps $5$s and $6$s, and re-rolls $1$s, $2$s, $3$s, and $4$s. We analyse how the agent’s predicted expected return in each state changes as one or both dice become known. These returns define the characteristic values in SVERL’s prediction game. The resulting Shapley values quantify each die’s contribution, as shown in \cref{tab:dicevalueestimation}.

\begin{figure}[t]
    \begin{center}
    \centerline{\includegraphics[width=0.35\columnwidth]{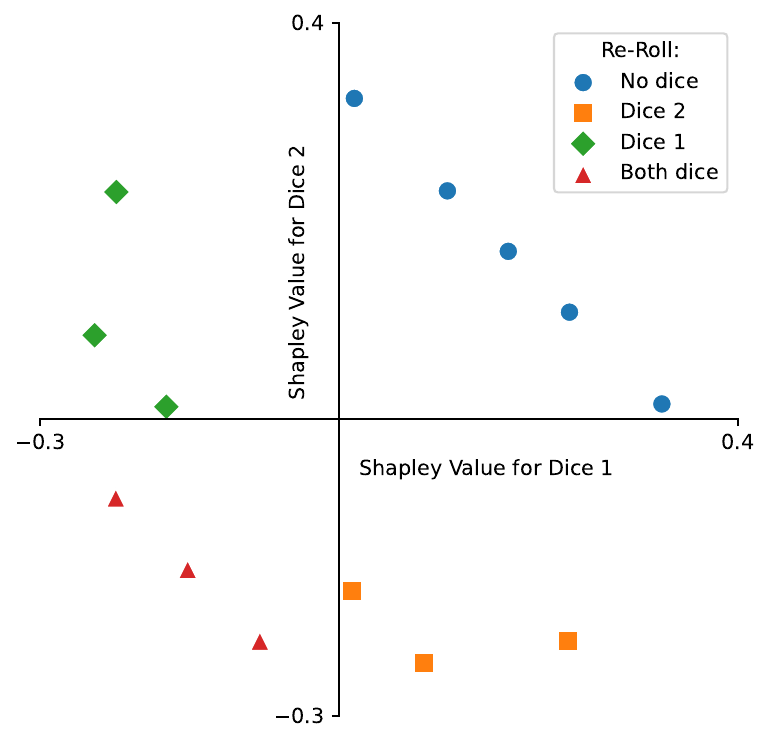}}
    \caption{Shapley values for the value estimates of every state in the dice-rolling game. Each point corresponds to a state, with the horizontal and vertical axes showing the Shapley value for the first and second die, respectively. Points are coloured according to which dice the optimal policy re-rolls.}
    \vspace{-2.5em}
    \label{fig:dice3_sverl_v}
    \end{center}
\end{figure}

Consider state $(3, 6)$. When neither die is known, the value estimate is the average return over the steady-state distribution:~$0.66$. If $d_2$ becomes known to be $6$, the estimate increases to $0.90$ because states with a $6$ have higher average returns. If instead $d_1$ becomes known to be $3$, the estimate drops to $0.45$ because states with a $3$ have lower average returns. The Shapley values reflect this difference: the $6$ contributes positively to the value estimate, while the $3$ contributes negatively. Now consider state $(1, 1)$. As above, when neither die is known, the estimate is $0.66$. If either die becomes known to be $1$, the estimate decreases to $0.45$ because states with a $1$ have a lower average return. When both dice are known, the agent must be in the lowest-return state, and the estimate falls further to $0.36$. Thus, each die lowers the value estimate by the same amount, as reflected in their equal negative Shapley values. Interestingly, this pattern holds across the entire state space. \cref{fig:dice3_sverl_v} shows the Shapley values for each die in every state, with each point coloured and shaped according to which dice are re-rolled under the optimal policy. The quadrant structure reveals a perfect correspondence: a die receives a negative attribution if and only if it is re-rolled. This illustrates how prediction-based explanations could inform control, highlighting features that should be changed to improve predicted return.

\begin{figure}[!t]
\begin{center}
\raisebox{-0.65\height}{\includegraphics[width=5.3in]{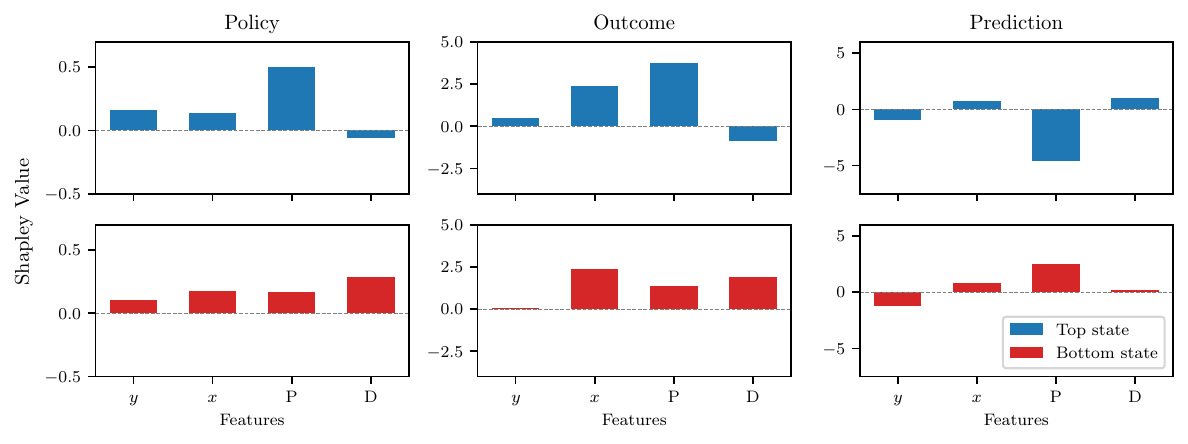}}
\raisebox{-0.68\height}{\includegraphics[width=0.5in]{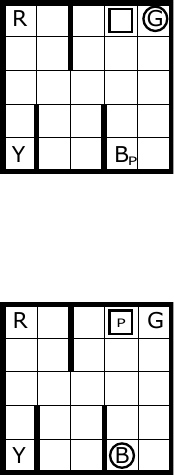}}
\caption{Shapley values for two states in \texttt{Taxi} (top and bottom), showing the feature contributions to the probability of selecting the optimal action taken (first column), to the agent’s expected return (second column), and the agent’s estimated expected return (third column). On the right-hand side, the corresponding taxi state is shown, with the taxi location marked by a rectangle, the passenger location with a \texttt{p}, and the destination with a circle.}
\label{fig:taxi_sverlp_sverlv_sverlpi}
\end{center}
\end{figure}

\textbf{Explaining Behaviour, Outcomes and Predictions.} We use the classic \texttt{Taxi} domain~\citep{Dietterich2000} to show how the same features influence each explanation differently depending on their role in the task and the agent’s progress toward its objective. In this domain, an agent navigates a gridworld, picking up a passenger and delivering them to their destination. Each state is described by four features: the taxi's $(x, y)$ position, the passenger location (\texttt{R}, \texttt{G}, \texttt{B}, \texttt{Y}, or \texttt{in-taxi}), and the destination (\texttt{R}, \texttt{G}, \texttt{B}, or \texttt{Y}). Each step incurs a reward of $-1$, with a bonus of $+20$ for a successful drop-off and a penalty of $-10$ for inappropriate pick-up and drop-off actions. We use the OpenAI Gym Taxi-v3 implementation~\citep{Brockman2016}. We analyse an optimal policy at two states (\cref{fig:taxi_sverlp_sverlv_sverlpi}).

In the top state, the taxi moves south towards the passenger at location \texttt{B}, and the destination is location \texttt{G}. The behaviour explanation highlights the passenger location as the dominant contributor, with the $x$ and $y$ coordinates contributing similarly. Surprisingly, the destination receives a small negative attribution, possibly because knowing it without knowing the passenger's location can increase the probability of taking suboptimal actions towards it. For outcome, the $x$ and $y$ coordinates contribute positively, though $x$ more than $y$, illustrating that while these features are similarly important for selecting actions, their consequences for return may differ. For prediction, the $y$ coordinate and passenger location lower the predicted expected return, reflecting that the agent is not in the same row as the passenger and has not picked them up.

In the bottom state, the taxi moves south to deliver the passenger, now onboard, to the destination at location \texttt{B}. The behaviour explanation now highlights the passenger and destination locations as similar key contributors. This illustrates the agent's competing goals: keep the passenger in the taxi and drop them off at the destination. The outcome explanation mirrors this pattern. In contrast to the top state, the passenger's presence in the taxi---indicating a higher return---now increases the predicted expected return.

%% file: plots/gwb.pdf_tex
\begingroup%
  \makeatletter%
  \providecommand\color[2][]{%
    \errmessage{(Inkscape) Color is used for the text in Inkscape, but the package 'color.sty' is not loaded}%
    \renewcommand\color[2][]{}%
  }%
  \providecommand\transparent[1]{%
    \errmessage{(Inkscape) Transparency is used (non-zero) for the text in Inkscape, but the package 'transparent.sty' is not loaded}%
    \renewcommand\transparent[1]{}%
  }%
  \providecommand\rotatebox[2]{#2}%
  \newcommand*\fsize{\dimexpr\f@size pt\relax}%
  \newcommand*\lineheight[1]{\fontsize{\fsize}{#1\fsize}\selectfont}%
  \ifx\svgwidth\undefined%
    \setlength{\unitlength}{82.8218841bp}%
    \ifx\svgscale\undefined%
      \relax%
    \else%
      \setlength{\unitlength}{\unitlength * \real{\svgscale}}%
    \fi%
  \else%
    \setlength{\unitlength}{\svgwidth}%
  \fi%
  \global\let\svgwidth\undefined%
  \global\let\svgscale\undefined%
  \makeatother%
  \begin{picture}(1,1.70513609)%
    \lineheight{1}%
    \setlength\tabcolsep{0pt}%
    \put(0,0){\includegraphics[width=\unitlength,page=1]{gwb.pdf}}%
    \put(0.42072956,0.4219981){\color[rgb]{0,0,0}\makebox(0,0)[t]{\lineheight{1.25}\smash{\begin{tabular}[t]{c}1\end{tabular}}}}%
    \put(0.75650567,0.42138512){\color[rgb]{0,0,0}\makebox(0,0)[t]{\lineheight{1.25}\smash{\begin{tabular}[t]{c}2\end{tabular}}}}%
    \put(0,0){\includegraphics[width=\unitlength,page=2]{gwb.pdf}}%
    \put(0.75685956,0.75713761){\color[rgb]{0,0,0}\makebox(0,0)[t]{\lineheight{1.25}\smash{\begin{tabular}[t]{c}3\end{tabular}}}}%
    \put(0,0){\includegraphics[width=\unitlength,page=3]{gwb.pdf}}%
    \put(0.4211543,1.09295685){\color[rgb]{0,0,0}\makebox(0,0)[t]{\lineheight{1.25}\smash{\begin{tabular}[t]{c}5\end{tabular}}}}%
    \put(0.75768475,1.09222579){\color[rgb]{0,0,0}\makebox(0,0)[t]{\lineheight{1.25}\smash{\begin{tabular}[t]{c}4\end{tabular}}}}%
    \put(0,0){\includegraphics[width=\unitlength,page=4]{gwb.pdf}}%
    \put(0.42287586,1.42729445){\color[rgb]{0,0,0}\makebox(0,0)[t]{\lineheight{1.25}\smash{\begin{tabular}[t]{c}G\end{tabular}}}}%
    \put(0.75789695,1.42729445){\color[rgb]{0,0,0}\makebox(0,0)[t]{\lineheight{1.25}\smash{\begin{tabular}[t]{c}G\end{tabular}}}}%
    \put(0,0){\includegraphics[width=\unitlength,page=5]{gwb.pdf}}%
    \put(0.5780829,0.16855352){\color[rgb]{0,0,0}\makebox(0,0)[lt]{\lineheight{1.25}\smash{\begin{tabular}[t]{l}$x$\end{tabular}}}}%
    \put(0,0){\includegraphics[width=\unitlength,page=6]{gwb.pdf}}%
    \put(0.19537206,0.61086986){\color[rgb]{0,0,0}\rotatebox{90}{\makebox(0,0)[lt]{\lineheight{1.25}\smash{\begin{tabular}[t]{l}$y$\end{tabular}}}}}%
    \put(0,0){\includegraphics[width=\unitlength,page=7]{gwb.pdf}}%
  \end{picture}%
\endgroup%